\documentclass[letterpaper]{article} 
\usepackage{aaai2026}
\usepackage{times}
\usepackage{helvet}
\usepackage{courier}
\usepackage[hyphens]{url}
\usepackage{graphicx} 
\usepackage{natbib} 
\usepackage{caption}
\usepackage{algorithm}

\usepackage{newfloat}
\usepackage{listings}
\DeclareCaptionStyle{ruled}{labelfont=normalfont,labelsep=colon,strut=off}
\floatstyle{ruled}
\newfloat{listing}{tb}{lst}{}
\floatname{listing}{Listing}

\usepackage{multirow}

\usepackage[table]{xcolor}

\usepackage{tikz}
\usepackage{comment}
\usepackage{color}

\usepackage{enumitem}
\usepackage{graphicx}
\usepackage{amsmath}
\usepackage{booktabs}
\usepackage{mathtools}
\usepackage{url}
\usepackage{comment}
\usepackage{color}
\usepackage{graphicx}
\usepackage{multirow, array, booktabs,makecell,xspace,bm}
\usepackage{pifont}
\usepackage{amsfonts}
\usepackage{arydshln}
\usepackage{dsfont}
\usepackage{xcolor}
\usepackage{colortbl}

\usepackage{stackengine}

\usepackage{scalerel}
\usepackage{esdiff} 

\usepackage{makecell}

\usepackage[x11names]{xcolor}
\usepackage{soul}
\sethlcolor{pink}

\definecolor{lightblue}{RGB}{217, 241, 252}
\definecolor{lightpink}{RGB}{252, 216, 215}
\newcommand{\hlred}[1]{\colorbox{lightpink}{\textcolor{red}{#1}}}

\newcommand{\hlblue}[1]{\colorbox{lightblue}{\textcolor{blue}{#1}}}

\usepackage{xcolor}

\usepackage{algorithm}
\usepackage{algpseudocode}
\algnewcommand{\LineComment}[1]{\State \# #1} 

\newcommand{\cmark}{\ding{51}}
\newcommand{\xmark}{\ding{55}}

\usepackage[most]{tcolorbox}
\tcbuselibrary{skins, breakable, listings}

\newtcolorbox{mybox}[1][]{
    enhanced,
    top=1pt,
    bottom=1pt,
    boxrule=0pt,
    arc=0pt,
    colframe=white,
    colback=white,
    borderline north={0.5mm}{0mm}{black},
    borderline south={0.5mm}{0mm}{black},
    #1
}
\newtcolorbox{mybox2}[1][]{
    enhanced,
    top=1pt,
    bottom=1pt,
    boxrule=0pt,
    arc=0pt,
    colframe=white,
    colback=white,
    borderline north={0.5mm}{0mm}{black},
    borderline south={0.5mm}{0mm}{black},
    width=\textwidth,
    left=0pt,
    right=0pt,
    #1
}

\newcommand{\supplementarytitle}[2][]{
    \twocolumn[
        \begin{center}
            {\LARGE \bfseries #2 \par}
            \vspace{0.5em}
            {\LARGE \itshape -- \textbf{Supplementary Material} -- \par}
            \vspace{2em}
        \end{center}
    ]
}

\usepackage{newfloat}
\usepackage{listings}
\DeclareCaptionStyle{ruled}{labelfont=normalfont,labelsep=colon,strut=off}
\floatstyle{ruled}
\newfloat{listing}{tb}{lst}{}
\floatname{listing}{Listing}

\title{Leveraging Textual Compositional Reasoning for Robust Change Captioning}
\author{
    Kyu Ri Park\textsuperscript{\rm 1}\equalcontrib,
    Jiyoung Park\textsuperscript{\rm 1}$\footnotemark[1]$,
    Seong Tae Kim\textsuperscript{\rm 1},
    Hong Joo Lee\textsuperscript{\rm 2}\thanks{Corresponding author}, 
    Jung Uk Kim\textsuperscript{\rm 1}$\footnotemark[2]$
}

\affiliations{
    \textsuperscript{\rm 1}Kyung Hee University, Yong-in, South Korea\\
    \textsuperscript{\rm 2}Technical University of Munich, Munich, Germany\\
    \{kyuri0924, jy0117, st.kim, ju.kim\}@khu.ac.kr, hongjoo.lee@tum.de
}

\usepackage{bibentry}

\begin{document}

\maketitle

\begin{abstract}

Change captioning aims to describe changes between a pair of images. However, existing works rely on visual features alone, which often fail to capture subtle but meaningful changes because they lack the ability to represent explicitly structured information such as object relationships and compositional semantics. To alleviate this, we present \textbf{CORTEX} (\textbf{CO}mpositional \textbf{R}easoning-aware \textbf{TEX}t-guided), a novel framework that integrates complementary textual cues to enhance change understanding. In addition to capturing cues from pixel-level differences, CORTEX utilizes scene-level textual knowledge provided by Vision Language Models (VLMs) to extract richer image text signals that reveal underlying compositional reasoning. CORTEX consists of three key modules: (\textit{i}) an Image-level Change Detector that identifies low-level visual differences between paired images, (\textit{ii}) a Reasoning-aware Text Extraction (RTE) module that use VLMs to generate compositional reasoning descriptions implicit in visual features, and (\textit{iii}) an Image-Text Dual Alignment (ITDA) module that aligns visual and textual features for fine-grained relational reasoning. This enables CORTEX to reason over visual and textual features and capture changes that are otherwise ambiguous in visual features alone. 
The code is available at \url{https://github.com/VisualAIKHU/CORTEX}.
\end{abstract}

\section{Introduction}
\label{sec:intro}

Change captioning aims to generate natural language descriptions that explain the differences between two images captured at different time points \cite{ spotthediff}. Unlike traditional image captioning, which focuses on describing a single static image, change captioning requires models to detect and clearly describe meaningful differences between image pairs. This is especially important in applications where capturing fine-grained visual differences is critical, such as in surveillance \cite{surveillance2} and medical imaging \cite{medicalRef}.

\begin{figure}[t]
    \begin{minipage}[b]{1.0\linewidth}
	\centering
        \centerline{\includegraphics[width=8.3cm]{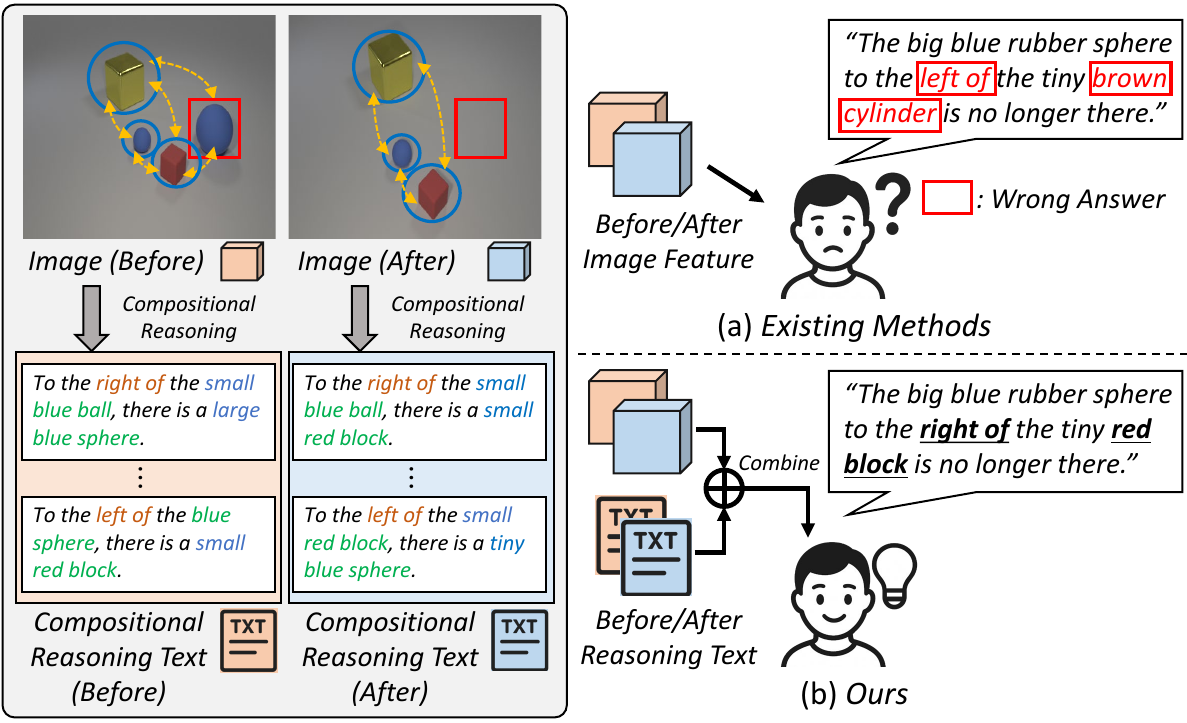}}
        \end{minipage}
	\caption{(a) Existing methods struggle to estimate changes because compositional reasoning cues are not explicitly represented in the image (\textit{i.e.,} object relationships (yellow arrows), spatial arrangements (blue circles)). (b) In contrast, our method incorporates explicit textual compositional reasoning cues to enhance scene understanding, thereby enabling more accurate change description.}
    \label{fig:1}
\end{figure}

Since it holds significant potential for various practical applications, extensive research has been conducted to develop algorithms that more effectively describe changes. SCORER \cite{scorer} introduces self-supervised cross-view contrastive alignment and cross-modal backward reasoning, which solely relies on visual features to distinguish true semantic changes from pseudo changes caused by viewpoint shifts. SMART \cite{smart} employs a multi-aspect relation learning network with a POS-based visual switch, relying on image data to capture semantic and positional relations. In addition, DIRL \cite{dirl} uses self-supervised channel correlation and decorrelation to align captions with visual differences, effectively addressing pseudo changes from viewpoint and illumination variations.

Despite these advances, existing methods adopt a visual-only approach, relying on image-level features to describe changes. While these features can capture low-level appearance differences, they often fail to support \textit{compositional reasoning}$-$the ability to understand structured semantics such as object relationships and spatial configurations. Since this type of information is not directly encoded in images but rather implicitly embedded \cite{text2scene}, models often struggle to generate accurate descriptions of changes. For example, as shown in Figure \ref{fig:1} (a), existing methods often misinterpret spatial relations (`\textit{left of}') or misidentify reference objects (`\textit{a tiny brown cylinder}'). These limitations emphasize the need for a change captioning model that can reason over compositional structures.

To address these limitations, we aim to enhance existing visual-only methods through a simple yet effective strategy that incorporates explicit textual cues conveying compositional reasoning. Unlike visual information, text can explicitly depict the structured semantics embedded in an image in a clear and interpretable form, serving as a strong signal for high-level reasoning \cite{text2scene}. As shown in Figure \ref{fig:1} (b), incorporating such compositional cues enables our model to better explain changes by capturing their relational and contextual meanings.

Based on the aforementioned points, we propose a novel \textbf{CO}mpositional \textbf{R}easoning-aware \textbf{TEX}t-guided (CORTEX) framework for robust change captioning. It consists of three modules. (\textit{i}) We employ an \textbf{image-level change detector} from visual-only approaches \cite{dirl, scorer, smart} to identify visual differences between the input image pairs. (\textit{ii}) To incorporate compositional understanding, we introduce a \textbf{Reasoning-aware Text Extraction (RTE) module}, which extracts explicit textual compositional reasoning cues (\textit{e.g.}, object relationships and spatial configurations). To this end, we leverage a widely adopted Vision-Language Model (VLM) with structured prompts to generate compositional descriptions for each image, offering rich semantic cues that enhance scene understanding. Also, (\textit{iii}) we propose an \textbf{Image-Text Dual Alignment (ITDA) module} that aligns visual and textual features via static alignment (within the scene) and dynamic alignment (cross scene). This dual alignment allows the model to embed the textual compositional reasoning into the visual features while preserving their representational strengths, leading to richer scene representations and improved structural change reasoning.

As a result, CORTEX generates more accurate captions by effectively describing changes between two images. Notably, the core objective of this work is to overcome the limitations of visual-only change captioning methods by introducing two \textit{plug-and-play} modules, namely RTE and ITDA, which can be seamlessly integrated into existing image-level change detectors to enhance compositional reasoning.

The major contributions of our paper are as follows:
\begin{itemize}
    \item We devise CORTEX, a new plug-and-play framework that enhances the existing visual-only approaches by incorporating explicit textual compositional reasoning, which was previously embedded implicitly in images and challenging for models to accurately infer.   
    \item We propose RTE module that leverages a VLM to extract structured textual cues that encode explicit compositional reasoning elements from images. The extracted text will be publicly released to facilitate future research.
    \item We introduce ITDA module, which aligns image and text features through static (within the scene) and dynamic (cross scene) alignment strategies to enable a more comprehensive understanding of both individual scenes and their differences.
\end{itemize}

\begin{figure*}[t]
    \begin{minipage}[b]{0.999\linewidth}
	\centering
        \centerline{\includegraphics[width=\linewidth]{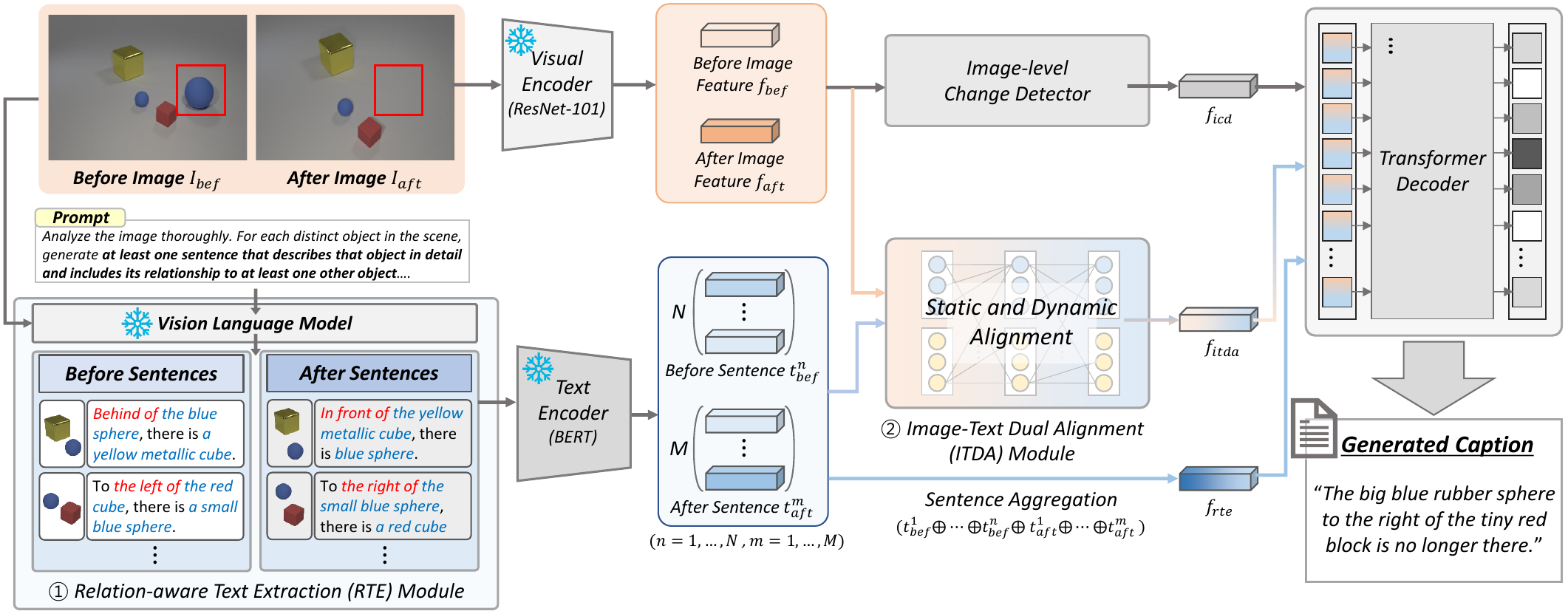}}
        \end{minipage}
    \centering
	\caption{Overview of the proposed Compositional Reasoning-aware Text-guided (CORTEX) framework for change captioning, which combines the three modules. We introduce (1) Image-level change detector, which captures change cues between the two images; (2) RTE module, which extracts compositional reasoning sentence for each scene; and (3) ITDA module, which reinforces same-scene understanding for static alignment and identifies changes in dynamic alignment in cross-scene.}
    \label{fig:2}
\end{figure*}

\section{Related Works}
\label{sec:RelatedWorks}
\subsection{Change Captioning}

Change captioning aims to generate natural language descriptions that capture semantic differences between two images, while filtering out irrelevant variations such as viewpoint or illumination changes. Earlier methods address irrelevant variations using techniques such as view-invariant representation learning, semantic alignment, and cross-modal regularization \cite{NCT, scorer, smart, dirl}.

Recent methods aim to suppress irrelevant variations and enhance cross-modal alignment for more accurate change descriptions. RDD \cite{rdd} mitigates noise from global difference features and improves linguistic-visual consistency through region-aware difference distillation and attribute-guided contrastive learning. DECIDER \cite{decider} addresses the limitations of auto-regressive decoders, such as error accumulation and weak inter-modality interaction. It adopts a contrastive diffusion framework with adversarial perturbations to generate more robust and semantically accurate change captions.

All visual-only methods struggle to capture fine-grained compositional and relational dynamics between objects. Our method addresses this gap by introducing textual compositionality into the training process, enabling more accurate and context-aware change descriptions.

\subsection{Vision-Language Models in Vision Tasks}
Vision-Language Models (VLMs) integrate visual and textual data to enable comprehensive image-text understanding \cite{vlm1}. As a result, they have been successfully applied to various vision tasks—including image classification and retrieval \cite{clip}, VQA \cite{vlm4}, image captioning \cite{vlm3}, object detection \cite{vlm_od}, and segmentation \cite{vlm_seg}—benefiting applications in autonomous driving \cite{vlm_autonomous} and medical imaging \cite{vlm_medical}.

While VLMs bridge visual and textual domains, our work leverages their strength in extracting relational and contextual cues from single images. Trained on large-scale image–text data, VLMs accurately capture semantics and object layouts \cite{vlm_multi_img}. However, optimized for single-image captioning, they struggle to compare two images and detect subtle relational nuances, as noted in prior analyses \cite{vlm_direct}.

Motivated by these limitations, our method leverages VLMs to extract relational context from individual images, supplementing visual-only change detectors. Instead of directly comparing paired images, we generate reasoning-aware text as auxiliary information. This fusion of high-level relational cues with low-level visual differences enables our approach to capture subtle change details and overall scene dynamics, leading to more accurate change captions.

\section{Methodology}
\label{sec:Method}
Figure \ref{fig:2} shows an overall architecture of CORTEX, a plug-and-play framework designed to enhance visual-only change captioning models by incorporating explicit compositional reasoning signals in textual form. CORTEX is composed of three modular components: (1) Image-level Change Detector, (2) Reasoning-aware Text Extraction (RTE) module, and (3) Image-Text Dual Alignment (ITDA) module. Given a pair of input images, \textit{``before"} image $I_{bef}$ and \textit{``after"} image $I_{aft}$, a visual backbone (\textit{e.g.,} ResNet-101 \cite{ResNet101}) extracts the corresponding visual features $f_{bef}$, $f_{aft}$. 
Then, the image-level change detector takes them as inputs and encodes $f_{icd}$, which captures low-level appearance differences between the two images. 
In parallel, the RTE module utilizes a VLM (\textit{e.g.,} InternVL2 \cite{internvl})  to extract $N$ compositional reasoning sentences ${T}_{bef}$ from $I_{bef}$ and $M$ sentences ${T}_{aft}$ from $I_{aft}$.
Each sentence contains high-level compositional reasoning cues, such as relative attributes (\textit{e.g.,} comparisons of size or brightness) and spatial relations that are difficult to infer from pixel-level signal alone.
A text encoder (\textit{e.g.,} BERT \cite{bert}) embeds $T_{bef}$, $T_{aft}$ into sentence features ${t}_{bef}$, ${t}_{aft}$. All the sentence features are concatenated to generate the RTE feature $f_{rte}$.
The ITDA module aligns visual features with the textual cues at both within and across scenes, generating the text-augmented feature $f_{itda}$ to support reasoning-aware scene understanding. The combination of compositional reasoning cues from RTE-generated texts and pixel-level visual features enables the model to effectively capture scene changes by complementing visual understanding with explicit textual reasoning. Finally, a transformer decoder generates change captions by integrating the outputs from all modules. More details are in the following subsections.

\subsection{Reasoning-aware Text Extraction (RTE) Module}
\label{RTE}
While existing image-level change detectors effectively capture appearance differences between two images, they often lack the ability to perform fine-grained contextual reasoning based on relative attributes and spatial context. To address this limitation, we introduce the RTE module, which extracts structured, sentence-level descriptions specially designed to support compositional reasoning.

In the RTE module, we leverage a frozen VLM (\textit{e.g.,} InternVL2 \cite{internvl}) to generate textual descriptions from $I_{bef}$ and $I_{aft}$.

Rather than generating generic descriptions, we prompt the VLM to extract compositional reasoning cues that include semantic details, which are often missed by visual features alone. To extract high-quality compositional-reasoning sentences, we use the following prompt:

\begin{tcolorbox}[colback=gray!5!white, colframe=gray!75!black, title=Prompt for Compositional Reasoning]
\textit{Analyze the image thoroughly. For each distinct object in the scene, generate \textbf{at least one sentence} that describes that object in detail and includes its relationship to at least one other object. Each sentence should mention \textbf{the object's color, shape, size, and relevant spatial relationships} (such as distance, proximity, or grouping).}
\end{tcolorbox}

This carefully crafted prompt encourages the generation of compositional reasoning cues by guiding the model to describe each object with detailed attributes (\textit{e.g.,} color, shape, size) and its spatial relationships \cite{vlm_compositional_reasoning} with other objects. It also ensures consistency and completeness, which are crucial for structured analysis.

At this time, since each scene varies in object density and complexity, the number of extracted sentences is determined dynamically based on the scene, denoted as  ${T}_{bef}{=}\{{T}_{bef}^{n}\}_{n=1}^{N}$, and  ${T}_{aft}{=}\{{T}_{aft}^{m}\}_{m=1}^{M}$ ($N$ sentences for `before' image and $M$ sentences for `after' image).

Subsequently, the generated sentences ${T}_{bef}$ and ${T}_{aft}$ are passed through a text encoder \cite{bert} to produce the sentence features ${t}_{bef}, {t}_{aft} \in \mathbb{R}^{c}$ ($c$ denotes the channel number). We will release all the generated sentences to support advanced research in change captioning. Detailed descriptions are provided in the supplementary materials.

\subsection{Image-Text Dual Alignment (ITDA) Module}
\label{ITDA}
Although the RTE module extracts compositional-reasoning text from individual images, these features are embedded in a different latent space than visual features from the image-level change detector. To unify image and text modalities and fully exploit their complementary strengths, we introduce the ITDA module. The ITDA module consists of two components: \textit{(i)} \textbf{static alignment}, which enhances compositional understanding within each scene, and \textit{(ii)} \textbf{dynamic alignment}, which emphasizes changes across scenes.\\

\begin{figure}[t]
    \begin{minipage}[b]{1.0\linewidth}
	\centering
        \centerline{\includegraphics[width=8.5cm]{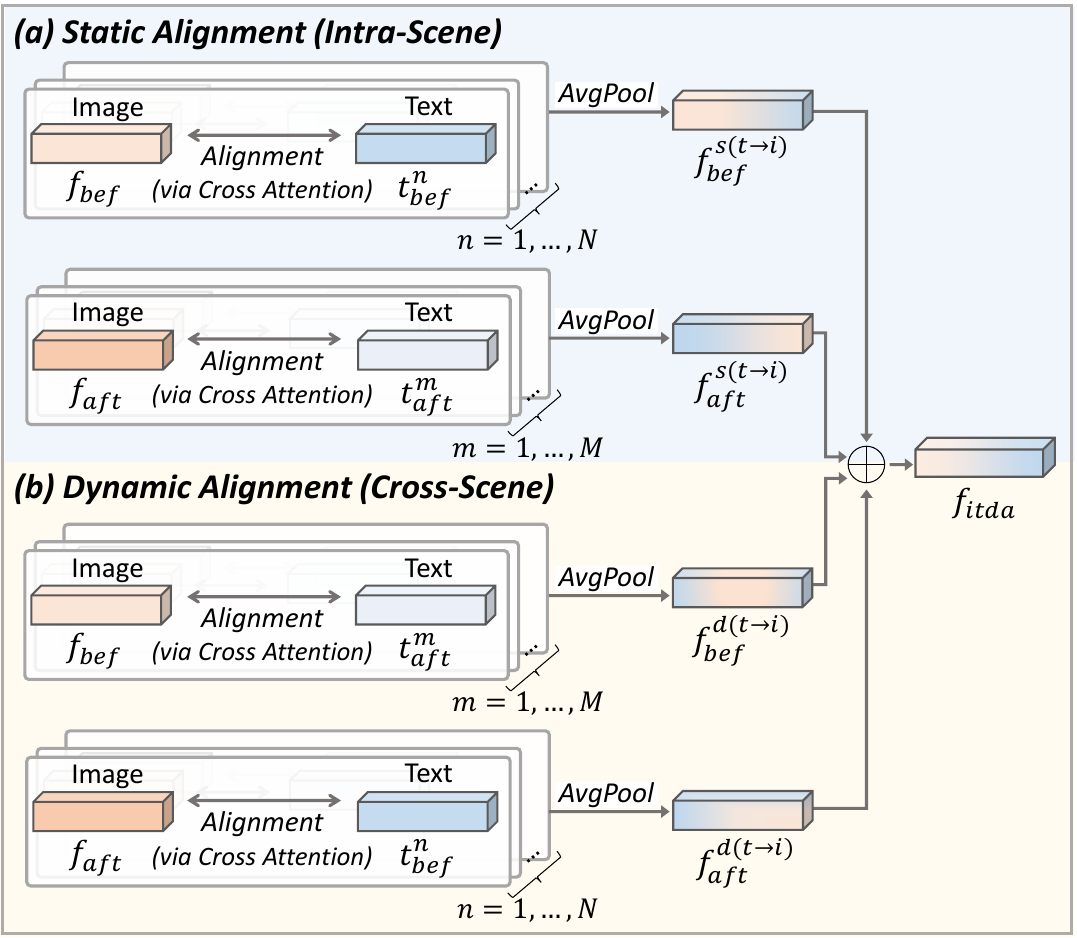}}
        \end{minipage}
	\caption{Overview of the ITDA module. (a) Static alignment matches each image with its corresponding compositional texts extracted by the RTE module. (b) Dynamic alignment matches each image with texts from the cross-scene to highlight the changes. $\bigoplus$ denotes concatenation.}
    \label{fig:4}
\end{figure}

\noindent\textbf{Static Alignment.} First, we design the static alignment to capture and refine intra-scene compositional structure. To do this, we align visual features with compositional-reasoning sentence features from the same scene (either \textit{``before"} or \textit{``after"}). As shown in Figure \ref{fig:4} (a), this process takes a visual feature $f_{bef}$ or $f_{aft}$ and the corresponding compositional-reasoning sentence feature $t_{bef}$ or $t_{aft}$. For the \textit{``before"} scene, $t_{bef}$ provides compositional reasoning cues that reflect relative attributes and spatial relationships. 
We apply cross-attention between visual feature $f_{bef}$ and each of the $N$ compositional-reasoning sentence features, then average the outputs, which can be formulated as:
\begin{equation}
    f_{bef}^{s(t\rightarrow i)} = \frac{1}{N} \sum_{n=1}^{N} \text{Attn}({t}_{bef}^{n}, f_{bef}, f_{bef}),
\end{equation}
where Attn($Q,K,V$) = Softmax$\left({Q K^\top}/{\sqrt{c}}\right) V$. This yields a text-augmented static feature for the \textit{``before"} image. The same process is applied to the \textit{``after" }scene, where its $M$ compositional-reasoning sentence features are used to produce the text-augmented static feature $f_{aft}^{s(t\rightarrow i)}$.

To ensure semantic consistency between $f_{bef}^{s(t\rightarrow i)}$, $f_{aft}^{s(t\rightarrow i)}$ and the image-level change detector feature $f_{icd}$, we compute self-attended visual features as follows:
\begin{table*}[t]
    \renewcommand{\tabcolsep}{3.8mm}
    \centering
    \resizebox{0.999\linewidth}{!}
    {
    \begin{tabular}{l ccccc ccccc}
        \Xhline{3\arrayrulewidth}
        \multicolumn{1}{c}{\multirow{2}{*}[-0.3em]{\bf Method}} & \multicolumn{5}{c}{\rule{0pt}{12pt}\bf Total Performance} & \multicolumn{5}{c}{\bf Semantic Change} \\
        \cmidrule(lr){2-6} \cmidrule(lr){7-11}
         & $\mathcal{B}$  & $\mathcal{M}$ & $\mathcal{R}$ & $\mathcal{C}$ & $\mathcal{S}$ & $\mathcal{B}$  & $\mathcal{M}$ & $\mathcal{R}$ & $\mathcal{C}$ & $\mathcal{S}$ \\
        \midrule
        DUDA (ICCV'19)
        & 47.3 & 33.9 & - & 112.3 & 24.5 & 42.9 & 29.7 & - & 94.6 & 19.9 \\
        DUDA+ (CVPR'21)
        & 51.2 & 37.7 & 70.5 & 115.4 & 31.1 & 49.9 & 34.3 & 65.4 & 101.3 & 27.9 \\
        MCCFormers-D (ICCV'21)
        & 52.4 & 38.3 & - & 121.6 & 26.8 & - & - & - & - & - \\
        MCCFormers-S (ICCV'21)
        & 57.4 & 41.2 & - & 125.5 & 32.4 & - & - & - & - & - \\
        PCL w/o Pre-training (AAAI'22)
        & 32.7 & 27.7 & 57.2 & 89.8 & - & - & - & - & - & - \\
        NCT (TMM'23)
        & 55.1 & 40.2 & 73.8 & 124.1 & 32.9 & 53.1 & 36.5 & 70.7 & 118.4 & 30.9 \\
        I3N-TD (TMM'23)
        & 55.8 & 40.6 & 73.9 & 125.6 & 32.8 & - & - & - & - & - \\
        VARD-Trans (TIP'23)
        & 55.4 & 40.1 & 73.8 & 126.4 & 32.6 & 53.6 & 36.7 & 71.0 & 119.1 & 30.5 \\
        RDD+ACR (AAAI'25) & 56.1 & 41.3 & 75.0 & 128.1 & 33.5 & - & - & - & - & - \\
        DECIDER (AAAI'25) & 56.4 & 39.7 & 75.3 & 131.3 & - & - & - & - & - & - \\\hline
        \rule{0pt}{10.5pt}SCORER (ICCV'23)
        & 56.3 & 41.2 & 74.5 & 126.8 & 33.3 & 54.4 & 37.6 & 71.7 & 122.4 & 31.6\\
        \cellcolor{gray!20}\textbf{CORTEX (SCORER)}
        &\cellcolor{gray!20}\textbf{57.0} & \cellcolor{gray!20}\textbf{42.7} & \cellcolor{gray!20}\textbf{75.9} & \cellcolor{gray!20}\textbf{128.8} & \cellcolor{gray!20}\textbf{33.9} & \cellcolor{gray!20}\textbf{54.9} & \cellcolor{gray!20}\textbf{39.2} & \cellcolor{gray!20}\textbf{74.0} & \cellcolor{gray!20}\textbf{127.5} & \cellcolor{gray!20}\textbf{32.8} \\\hline
        \rule{0pt}{10.5pt}SMART (TPAMI'24)
        & 56.1 & 40.8 & 74.2 & 127.0 & 33.4 & 54.3 & 37.4 & 71.8 & 123.6 & 32.0 \\
        \cellcolor{gray!20}\textbf{CORTEX (SMART)} %
        &\cellcolor{gray!20}\textbf{56.5} & \cellcolor{gray!20}\textbf{42.1} & \cellcolor{gray!20}\textbf{75.7} & \cellcolor{gray!20}\textbf{130.2} & \cellcolor{gray!20}\textbf{34.0} & \cellcolor{gray!20}\textbf{54.6} & \cellcolor{gray!20}\textbf{39.1} & \cellcolor{gray!20}\textbf{74.5} & \cellcolor{gray!20}\textbf{130.3} & \cellcolor{gray!20}\textbf{32.9} \\\hline
        \rule{0pt}{10.5pt}DIRL (ECCV'24)
        & - & - & - & - & - & 54.6 & 38.1 & 71.9 & 123.6 & 31.8 \\
        DIRL$^{\dagger}$ (ECCV'24)
        &{55.5} & {40.8} & {73.4} & {125.3} & {33.4} & \textbf{55.4} & {38.4} & {72.1} & {123.2} & {32.7}
        \rule{0pt}{5pt} \\
        \cellcolor{gray!20}\textbf{CORTEX (DIRL)}
        &\cellcolor{gray!20}\textbf{57.4} & \cellcolor{gray!20}\textbf{43.0} & \cellcolor{gray!20}\textbf{76.2} & \cellcolor{gray!20}\textbf{130.7} & \cellcolor{gray!20}\textbf{34.2} & \cellcolor{gray!20}\textbf{55.4} & \cellcolor{gray!20}\textbf{39.6} & \cellcolor{gray!20}\textbf{74.6} & \cellcolor{gray!20}\textbf{131.1} & \cellcolor{gray!20}\textbf{33.5}
        \rule{0pt}{5pt} \\
        \Xhline{3\arrayrulewidth}
    \end{tabular}
    }
    \caption{Performance comparisons on the CLEVR-Change dataset (BLEU-4 ($\mathcal{B}$), METEOR ($\mathcal{M}$), ROUGE-L ($\mathcal{R}$), CIDEr ($\mathcal{C}$), and SPICE ($\mathcal{S}$)). CORTEX consistently improves the performances of existing methods that provide publicly available source code. $^{\dagger}$ denotes reproduced results of the method, as DIRL does not report `total' performance.}
    \label{table:1}
\end{table*}


\begin{equation}
    f_{bef}^{s(i\rightarrow i)} = \text{Attn}(f_{bef},f_{bef},f_{bef}),
\end{equation}
\begin{equation}
    f_{aft}^{s(i\rightarrow i)} = \text{Attn}(f_{aft},f_{aft},f_{aft}).
\end{equation}
Then, we introduce a static alignment loss $\mathcal{L}_{sa}$ to encourage both latent spaces to share common semantics, denoted as:
\begin{equation}
\mathcal{L}_{sa} = \frac{1}{2}(\| f_{bef}^{s(t\rightarrow i)} - f_{bef}^{s(i\rightarrow i)} \|^2_2 + \| f_{aft}^{s(t\rightarrow i)} - f_{aft}^{s(i\rightarrow i)} \|^2_2).
\end{equation}

This loss guides the model to integrate compositional details into visual representations, thereby improving scene-level compositional reasoning through static alignment. \\

\noindent\textbf{Dynamic Alignment.} Further, to capture the cross-scenes changes, we devise the dynamic alignment by applying cross-attention between the visual features of one scene and the compositional-reasoning sentence features of the other. As shown in Figure \ref{fig:4} (b), for the \textit{``before"} scene, we use its visual feature $f_{bef}$ with $M$ \textit{``after"} sentence feature $t_{aft}$. Then the outputs are averaged to produce the final text-augmented dynamic feature for the \textit{``before"} scene:
\begin{equation}
    f_{bef}^{d(t\rightarrow i)} = \frac{1}{M} \sum_{m=1}^{M} \text{Attn}({t}_{aft}^m, f_{bef}, f_{bef}),
\end{equation}
and \textit{``after"} text-augmented dynamic feature $f_{aft}^{d(t\rightarrow i)}$ is computed through a similar process.
To ensure that these text-augmented dynamic features align with the visual features, we design the dynamic alignment loss $\mathcal{L}_{da}$ as: 
\begin{equation}
    f_{bef}^{d(i\rightarrow i)} = \text{Attn}(f_{aft},f_{bef},f_{bef}),
\end{equation}
\begin{equation}
    f_{aft}^{d(i\rightarrow i)} = \text{Attn}(f_{bef},f_{aft},f_{aft}),
\end{equation}
\begin{equation}
\mathcal{L}_{da} = \frac{1}{2}(\| f_{bef}^{d(t\rightarrow i)} - f_{bef}^{d(i\rightarrow i)} \|^2_2 + \| f_{aft}^{d(t\rightarrow i)} - f_{aft}^{d(i\rightarrow i)} \|^2_2),
\end{equation}
where $f^{d(i\rightarrow i)}$ denotes cross-attended visual features. 

The $\mathcal{L}_{da}$ enforces alignment between the visual features from one scene and the text feature of another scene, facilitating more effective identification of scene differences by leveraging complementary multimodal information.

Finally, $f_{bef}^{s(t\rightarrow i)}$, $f_{aft}^{s(t\rightarrow i)}$, $f_{bef}^{d(t\rightarrow i)}$ and $f_{aft}^{d(t\rightarrow i)}$ are concatenated to generate the final feature $f_{itda}$.
We devise alignment loss $\mathcal{L}_{align}$ which combines $\mathcal{L}_{sa}$ and $\mathcal{L}_{da}$, denoted as:
\begin{equation}
    \mathcal{L}_{align} = \mathcal{L}_{sa} + \mathcal{L}_{da}.
\end{equation}

As a result, ITDA module enables a unified understanding of both intra-scene composition and cross-scene differences by leveraging compositional-reasoning sentences. This dual alignment compensates for the limitation of visual-only approaches by providing semantically enriched and contextually grounded representations.

Building on this alignment, we define the total training objective of CORTEX as follows:
\begin{equation}
    \mathcal{L}_{total} = \mathcal{L}_{cap} + \lambda\mathcal{L}_{align},
\end{equation}
where $\lambda$ is balancing hyper-parameter, $\mathcal{L}_{cap}$ represents the captioning loss that guides the overall caption generation process of the transformer decoder \cite{dirl}.

\section{Experiments}
\label{sec:Experiments}
\subsection{Datasets and Evaluation Metrics}
\noindent\textbf{Datasets.} In our experiments, we use three datasets: (\textit{i}) CLEVR-Change, (\textit{ii}) CLEVR-DC, and (\textit{iii}) Spot-the-Diff. 
\textbf{CLEVR-Change} \cite{DUDA} is a large-scale synthetic dataset for controlled settings, consisting of 79,606 image pairs and 493,735 captions. It is divided into 67,660 training, 3,976 validation, and 7,970 test pairs.

\textbf{CLEVR-DC} \cite{clevrdc} contains 48,000 image pairs with additional dynamic viewpoint shifts as distractors. We follow the official split: 85\% training, 5\% validation, and 10\% testing.

\textbf{Spot-the-Diff} \cite{spotthediff} consists of 13,192 well-aligned surveillance image pairs, divided into 8:1:1 splits for training, validation, and testing, providing a real-world benchmark for generalization. \\

\noindent\textbf{Evaluation Metrics.} We evaluate caption quality using standard metrics: BLEU-4 ($\mathcal{B}$) \cite{bleu}, METEOR ($\mathcal{M}$) \cite{meteor}, ROUGE-L ($\mathcal{R}$) \cite{rouge}, CIDEr ($\mathcal{C}$) \cite{cider}, and SPICE ($\mathcal{S}$) \cite{spice}.

\begin{table}[t]
    \renewcommand{\tabcolsep}{2.3mm}
    \centering
    \small
    \begin{tabular}{lccccc}
        \Xhline{3\arrayrulewidth}
        \multicolumn{1}{c}{\rule{0pt}{11pt}\textbf{Method}} & $\mathcal{B}$  & $\mathcal{M}$ & $\mathcal{R}$ & $\mathcal{C}$ & $\mathcal{S}$ \\
        \midrule
        DUDA (ICCV'19)
        & 40.3 & 27.1 & - & 56.7 & 16.1 \\
        M-VAM (ECCV'20)
        & 40.9 & 27.1 & - & 60.1 & 15.8 \\
        VACC (ICCV'21)
        & 45.0 & 29.3 & - & 71.7 & 17.6 \\
        NCT (TMM'23)
        & 47.5 & 32.5 & 65.1 & 76.9 & 15.6 \\
        VARD-Trans (TIP'23)
        & 48.3 & 32.4 & - & 77.6 & 15.4 \\
        \hline
        \rule{0pt}{9pt}SCORER (ICCV'23)
        & 49.4 & 33.4 & 66.1 & 83.7 & 16.2 \\
        \cellcolor{gray!20}\textbf{CORTEX (SCORER)}
        & \cellcolor{gray!20}\textbf{52.2}
        & \cellcolor{gray!20}\textbf{34.0}
        & \cellcolor{gray!20}\textbf{67.3}
        & \cellcolor{gray!20}\textbf{88.7}
        & \cellcolor{gray!20}\textbf{16.5} \\
        \hline
        \rule{0pt}{9pt}SMART$^{\dagger}$ (TPAMI'24)
        & 47.9 & 32.7 & 65.4 & 82.9 & 15.6 \\
        \cellcolor{gray!20}\textbf{CORTEX (SMART)}
        & \cellcolor{gray!20}\textbf{53.2}
        & \cellcolor{gray!20}\textbf{32.9}
        & \cellcolor{gray!20}\textbf{66.6}
        & \cellcolor{gray!20}\textbf{86.6}
        & \cellcolor{gray!20}\textbf{16.9} \\
        \hline
        \rule{0pt}{9pt}DIRL (ECCV'24)
        & 51.4 & 32.3 & 66.3 & 84.1 & 16.8 \\
        \cellcolor{gray!20}\textbf{CORTEX (DIRL)}
        & \cellcolor{gray!20}\textbf{55.3}
        & \cellcolor{gray!20}\textbf{32.9}
        & \cellcolor{gray!20}\textbf{67.8}
        & \cellcolor{gray!20}\textbf{89.7}
        & \cellcolor{gray!20}\textbf{17.0} \\
        \Xhline{3\arrayrulewidth}
    \end{tabular}
    \caption{Performance comparisons with state-of-the-art methods on the CLEVR-DC dataset.}
    \label{table:2}
\end{table}

\subsection{Implementation Details}
We adopt three state-of-the-art baselines with publicly available code: SCORER, SMART, and DIRL \cite{scorer, smart, dirl} as our image-level change detectors. At this stage, following the standard protocol in change captioning, we use a pre-trained ResNet-101 \cite{ResNet101} to extract features from image pairs. We use InternVL2-8B \cite{internvl} as the VLM to generate compositional reasoning texts.

We set the number of attention heads to $h=8$ and $\lambda=10^{-3}$ for SCORER and $\lambda=10^{-4}$ for both SMART and DIRL in Eq. (10).
CORTEX is trained using Adam optimizer \cite{adam} on a single RTX 4090 GPU.

\subsection{Comparison with Existing Methods}
\noindent\textbf{Results on the CLEVR-Change Dataset.} We compared our method with SOTA methods on the CLEVR-Change dataset. We evaluated performance for both `total' and `semantic change' settings \cite{dirl}, where `semantic change' refers to cases with actual changes, and `total' includes both changed and unchanged cases.

As shown in Table \ref{table:1}, incorporating compositional reasoning through CORTEX into three SOTAs (SCORER, SMART, DIRL) consistently improves performance. This shows that CORTEX effectively complements visual-only methods by injecting explicit compositional reasoning into the change captioning process. The consistent gains across baselines confirm its effectiveness and the importance of compositional understanding. \\

\begin{table}[t!]
    \renewcommand{\tabcolsep}{1.8mm}
    \centering
    \small
    \begin{tabular}{lccccc}
        \Xhline{3\arrayrulewidth}
        \multicolumn{1}{c}{\rule{0pt}{11pt}\textbf{Method}} & $\mathcal{B}$  & $\mathcal{M}$ & $\mathcal{R}$ & $\mathcal{C}$ & $\mathcal{S}$ \\
        \midrule
        DUDA+ (CVPR'21)       & 8.1 & 12.5 & -    & 34.5 & -    \\
        MCCFormers-D (ICCV'21)&10.0 & 12.4 & -    & 43.1 & 18.3 \\
        MCCFormers-S (ICCV'21)& -   & 12.3 & -    & 41.6 & 16.3 \\
        I3N-TD (TMM'23)       & -   & 13.0 & 31.5 & 42.7 & 18.6 \\
        VARD-Trans (TIP'23)   & -   & 12.5 & 29.3 & 30.3 & 17.3 \\
        RDD+ACR (AAAI'25)     & 9.2 & 13.9 & 31.0 & 43.6 & -    \\
        DECIDER (AAAI'25)     &10.7 & 14.2 & 41.6 & 39.9 & -    \\
        \hline
        \rule{0pt}{8.5pt}SCORER (ICCV'23)      &10.2 & 12.2 & -    & 38.9 & 18.4 \\
        \cellcolor{gray!20}\textbf{CORTEX (SCORER)} &
        \cellcolor{gray!20}\textbf{10.5} &
        \cellcolor{gray!20}\textbf{12.6} &
        \cellcolor{gray!20}33.2 &
        \cellcolor{gray!20}\textbf{40.3} &
        \cellcolor{gray!20}\textbf{19.4} \\
        \hline
        \rule{0pt}{8.5pt}SMART (TPAMI'24)      & -   & \textbf{13.5} & 31.6 & 39.4 & \textbf{19.0} \\
        \cellcolor{gray!20}\textbf{CORTEX (SMART)} &
        \cellcolor{gray!20}9.5 &
        \cellcolor{gray!20}12.2 &
        \cellcolor{gray!20}\textbf{32.7} &
        \cellcolor{gray!20}\textbf{41.0} &
        \cellcolor{gray!20}\textbf{19.0} \\
        \hline
        \rule{0pt}{8.5pt}DIRL (ECCV'24) &10.3 & 13.8 & 32.8 & 40.9 & 19.9 \\
        \cellcolor{gray!20}\textbf{CORTEX (DIRL)} &
        \cellcolor{gray!20}\textbf{11.6} &
        \cellcolor{gray!20}\textbf{13.9} &
        \cellcolor{gray!20}\textbf{33.4} &
        \cellcolor{gray!20}\textbf{49.5} &
        \cellcolor{gray!20}\textbf{21.4} \\
        \Xhline{3\arrayrulewidth}
    \end{tabular}
    \caption{Performance comparisons with state-of-the-art methods on the Spot-the-Diff dataset.}
    \label{table:3}
\end{table}


 \begin{table}[t]
     \centering
     \renewcommand{\tabcolsep}{2.8mm}
     \resizebox{\linewidth}{!}
     {
    \begin{tabular}{cc ccccc}
         \Xhline{3\arrayrulewidth}
         \multirow{2}{*}[-0.8ex]{RTE} & \multirow{2}{*}[-0.8ex]{ITDA} & \multicolumn{5}{c}{\rule{0pt}{11pt}\bf Total} \\
         \cmidrule(lr){3-7}
         & & $\mathcal{B}$  & $\mathcal{M}$ & $\mathcal{R}$ & $\mathcal{C}$ & $\mathcal{S}$ \\
         \midrule
         \multicolumn{2}{c}{Baseline (DIRL)} & 55.5 & 40.8 & 73.4 & 125.3 & 33.4 \\\cdashline{1-7}
         \rule{0pt}{10.0pt}\cmark & - & 55.8 & 41.6 & 74.8 & 128.5 & 33.9 \\
         \cmark & \cmark & \textbf{57.4} & \textbf{43.0} & \textbf{76.2} & \textbf{130.7} & \textbf{34.2} \\
         \Xhline{3\arrayrulewidth}
     \end{tabular}
     }
     \caption{Effect of the proposed modules (RTE and ITDA) on the CLEVR-Change dataset.}
     \label{table:4}
 \end{table}

\noindent\textbf{Results on the CLEVR-DC Dataset.} 
We evaluated performance on the CLEVR-DC dataset to assess robustness under extreme viewpoint changes. As shown in Table \ref{table:2}, integrating CORTEX into three SOTA methods consistently achieved the best results across all metrics. By incorporating VLM-extracted textual cues, CORTEX enhances image-text understanding and enables robust compositional reasoning even under drastic viewpoint shifts.\\

 \begin{table}[t!]
     \centering
     \renewcommand{\tabcolsep}{3.5mm}
     \resizebox{\linewidth}{!}
     {
    \begin{tabular}{cc ccccc}
         \Xhline{3\arrayrulewidth}
         \multicolumn{2}{c}{\rule{0pt}{11pt} $\mathcal{L}_{align}$} & \multicolumn{5}{c}{\bf Total} \\
         \cmidrule(lr){1-2} \cmidrule(lr){3-7}
         $\mathcal{L}_{sa}$ & $\mathcal{L}_{da}$ & $\mathcal{B}$  & $\mathcal{M}$ & $\mathcal{R}$ & $\mathcal{C}$ & $\mathcal{S}$ \\
         \midrule
         \xmark & \xmark & 56.6 & 41.5 & 75.1 & 127.9 & 33.5 \\\cdashline{1-7}
         \rule{0pt}{10.0pt}\cmark & \xmark & 56.3 & 41.8 & 75.5 & 128.4 & 34.0 \\
         \xmark & \cmark & 56.6 & 41.8 & 75.6 & 128.9 & 33.7 \\
         \cmark & \cmark & \textbf{57.4} & \textbf{43.0} & \textbf{76.2} & \textbf{130.7} & \textbf{34.2} \\
         \Xhline{3\arrayrulewidth}
     \end{tabular}
     }
     \caption{Effect of the proposed static loss ($\mathcal{L}_{sa}$) and dynamic loss ($\mathcal{L}_{da}$) on the CLEVR-Change dataset.}
     \label{table:5}
 \end{table}


\begin{table}[t!]
    \renewcommand{\tabcolsep}{1.9mm}
    \centering
    \resizebox{0.999\linewidth}{!}
    {
    \begin{tabular}{c ccccc}
        \Xhline{3\arrayrulewidth}
        \rule{0pt}{11pt}\bf Prompt Type
         & $\mathcal{B}$  & $\mathcal{M}$ & $\mathcal{R}$ & $\mathcal{C}$ & $\mathcal{S}$\\
        \midrule
        Generic descriptions
        & 56.5 & {41.6} & {75.3} & {129.5} & {33.5}\\
        Compositional reasoning
        & \textbf{57.4} & \textbf{43.0} & \textbf{76.2} & \textbf{130.7} & \textbf{34.2} \\
        \Xhline{3\arrayrulewidth}
    \end{tabular}
    }
    \caption{Effect of different VLM prompt types on the CLEVR-Change dataset.}
    \label{table:6}
\end{table}

\begin{table}[t]
    \renewcommand{\tabcolsep}{1.5mm}
    \centering
    \resizebox{\linewidth}{!}
    {
    \begin{tabular}{cccccc}
        \Xhline{3\arrayrulewidth}
        \rule{0pt}{11pt}\bf VLM & $\mathcal{B}$  & $\mathcal{M}$ & $\mathcal{R}$ & $\mathcal{C}$ & $\mathcal{S}$ \\
        \midrule
        Baseline & 55.5 & 40.8 & 73.4 & 125.3 & 33.4 \\\cdashline{1-6}
        \rule{0pt}{11.5pt}
        LLaVA \cite{llava} & 56.9 & 42.3 & 75.7 & 130.0 & 34.1 \\
        InternVL2 \cite{internvl} & \textbf{57.4} & \textbf{43.0} & \textbf{76.2} & \textbf{130.7} & \textbf{34.2} \\
        \Xhline{3\arrayrulewidth}
    \end{tabular}
    }
    \caption{Comparison of different VLMs used in our method on the CLEVR-Change dataset. The top row shows the visual-only baseline (DIRL) without text-based guidance.}
    \label{table:7}
\end{table}

\begin{figure*}[t]
    \begin{minipage}[b]{0.86\linewidth}
	\centering
        \centerline{\includegraphics[width=\linewidth]{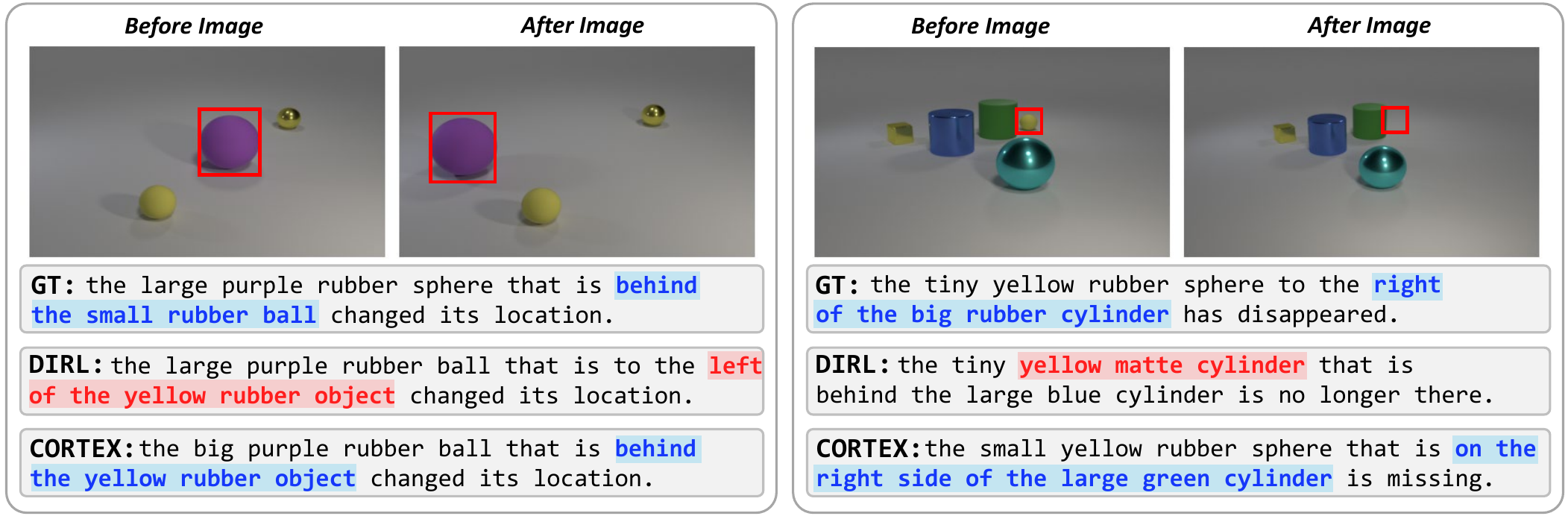}}
        \end{minipage}
    \centering
	\caption{Visualization examples in the CLEVR-Change dataset (\hlblue{Blue}/\hlred{red}: correct/incorrect compositional reasoning cues).}
    \label{fig:5}
\end{figure*}

\noindent\textbf{Results on the Spot-the-Diff Dataset.}
Further, to validate the generalization ability of our method in real-world scenes, we conducted experiments on the Spot-the-Diff dataset, which consists of image pairs from surveillance cameras. In Table \ref{table:3}, applying CORTEX yields superior performance across most metrics. These results emphasize the robustness of our architecture, showing its ability to generalize across diverse real-world scenes and change contexts.

\subsection{Ablation Studies}
We conducted ablation studies to analyze the effect of (\textit{i}) the proposed modules, (\textit{ii}) the proposed losses, and (\textit{iii}) different VLM prompt types. All variants were evaluated under the `total' setting on the CLEVR-Change dataset, using DIRL \cite{dirl} as the baseline. \\

\noindent\textbf{Effect of the Proposed Modules.} 
We evaluated the impact of the two proposed modules, RTE and ITDA. Table \ref{table:4} shows that applying the RTE module, which leverages compositional reasoning text significantly outperforms the visual-only baseline (DIRL). The best performance is achieved by using both modules. Note that ITDA was not tested alone, as it requires the compositional reasoning text generated by the RTE module. \\

\noindent\textbf{Effect of the Proposed Losses.}
Table \ref{table:5} shows the effect of the two losses, $\mathcal{L}_{sa}$ and $\mathcal{L}_{da}$, used in $\mathcal{L}_{align}$ of the ITDA module. Adding either loss individually led to improvements over the baseline without both losses. We achieved the highest performance when both losses were considered. \\

\noindent\textbf{Effect of Prompt Types for VLM.} 
To investigate the impact of reasoning cues on the visual-only method, we compare two types of VLM-generated text: (1) generic descriptions prompted with “\textit{Analyze the image and list sentences that describe the scene.}” and (2) compositional reasoning sentences guided to include relative attributes and inter-object relationships. In Table \ref{table:6}, compositional reasoning consistently yields better performance, highlighting the importance of structured cues for inferring fine-grained relational and attribute information essential to describing changes.

\subsection{Qualitative Results}
We compared CORTEX with DIRL in Figure \ref{fig:5}. CORTEX better captures compositional structures, generating outputs more closely aligned with the ground truth (see \hlblue{blue}/\hlred{red} parts). This demonstrates the strength of compositional reasoning, where textual guidance enables the model to produce more precise and context-aware change descriptions.

\subsection{Discussion}

\noindent\textbf{Effect of VLM Variants on CORTEX.}
While InternVL2 is our primary VLM, we also evaluated CORTEX with LLaVA \cite{llava}. As shown in Table \ref{table:7}, CORTEX consistently outperforms the baseline, effectively leveraging image-text understanding across different VLMs. \\

\begin{table}[t!]
    \renewcommand{\tabcolsep}{2.2mm}
    \centering
    \resizebox{\linewidth}{!}
    {
    \begin{tabular}{cccccc}
        \Xhline{3\arrayrulewidth}
        \rule{0pt}{9.8pt} \bf VLM Usage & $\mathcal{B}$  & $\mathcal{M}$ & $\mathcal{R}$ & $\mathcal{C}$ & $\mathcal{S}$ \\
        \midrule
        \addlinespace[1mm]
        \makecell{Direct Prediction \\ (VLM for \textit{Paired} Images)} & 2.7 & 10.7 & 21.0 & 12.3 & 12.5 \\
        \cdashline{1-6}
        \addlinespace[1mm]
        \makecell{Auxiliary Context (Ours) \\ (VLM for \textit{Single} Image)} & \textbf{11.6} & \textbf{13.9} & \textbf{33.4} & \textbf{49.5} & \textbf{21.4} \\
        \Xhline{3\arrayrulewidth}
    \end{tabular}
    }
    \caption{Comparison of VLM usage strategies on the Spot-the-Diff dataset. Our method enhances visual-only change captioning with compositional reasoning from single-image captions, outperforming direct paired-image approach.}
    \label{table:8}
\end{table}
 \begin{table}[t]
     \centering
     \renewcommand{\tabcolsep}{0.6mm}
     \resizebox{\linewidth}{!}
     {
    \begin{tabular}{c cccccc}
         \Xhline{3\arrayrulewidth}
         \addlinespace[1mm]
         \multirow{3}{*}{\rule{0pt}{1.5ex}\textbf{Method}} & 
         \multicolumn{1}{c}{\rule{0pt}{2ex}\bf Offline} &
        \multicolumn{3}{c}{\rule{0pt}{2ex}\bf Online} \\
        \cmidrule(lr){2-2} \cmidrule(lr){3-5} &  
        \makecell{\textbf{VLM} \\ (\textit{per img})} &  \makecell{\textbf{Train.} \\ \textit{(per iter)}} & \makecell{\textbf{Infer.} \\ (\textit{per img})} & \makecell{\textbf{\#Learnable} \\ \textbf{params}}\\
         \midrule
         DIRL (ECCV'24) & -  & 0.77s & 0.007s & 14.4M\\
         CORTEX (DIRL) & 3.94s & 0.79s & 0.008s & 18.2M\\
         \Xhline{3\arrayrulewidth}
     \end{tabular}
     }
     \caption{Offline/online time analysis with baseline DIRL.}
     \label{table:9}
 \end{table}

\noindent\textbf{Comparison of VLM Usage Strategies.} As VLMs are trained to understand semantics and spatial layouts from \textit{single images}, they often \textit{struggle to compare two images and detect subtle differences}, as noted in \cite{vlm_direct}.

To investigate this issue, we compared two VLM usage strategies on the Spot-the-Diff dataset (Table \ref{table:8}). The first approach directly infers changes by feeding paired ``\textit{before}” and ``\textit{after}” images into the VLM with a prompt to identify changes (in the supplementary). In contrast, second approach (Ours) leverages the single-image reasoning ability of VLMs by extracting compositional reasoning sentences from individual images and incorporating them as auxiliary cues into visual-only methods \cite{dirl}. These sentences provide fine-grained semantic cues (\textit{e.g.}, object attributes and spatial relations) often missed by visual-only models, enabling CORTEX to better capture subtle changes and compositional context. \\

\noindent\textbf{Computational Costs.}
Table \ref{table:9} compares offline captioning time, online training/inference times, and learnable parameters. Following \cite{computational_cost}, captions are generated with the VLM (InternVL2) offline before training, reducing runtime overhead. While VLM-based captioning adds one-time preprocessing step, its impact on efficiency is minimal, keeping our method lightweight and compositional-aware. \\

\noindent\textbf{Limitations.}
While our framework achieves strong performance by utilizing compositional reasoning cues from VLM-generated text, the use of VLMs incurs computational overhead. For more practical applications, our future work will aim to reduce this overhead.

\section{Conclusion}

We introduced CORTEX, a new framework for change captioning that incorporates textual compositional reasoning cues while preserving the strengths of image-level change detectors. To this end, we devise two modules: RTE for extracting compositional cues and ITDA for aligning image-text features to capture both static and dynamic changes. Designed as a plug-and-play component, CORTEX can be seamlessly integrated into existing methods. Experimental results highlight the importance of explicit compositional reasoning in accurately describing scene changes.

\section{Acknowledgments}

This work was supported by the NRF grant funded by the Korea government (MSIT) (No. RS-2023-00252391), and by IITP grant funded by the Korea government (MSIT) (No. RS-2025-25442384, IITP-2023-RS-2023-00266615: Convergence Security Core Talent Training Business Support Program, IITP-2022-II220078: Explainable Logical Reasoning for Medical Knowledge Generation) and conducted by CARAI grant funded by DAPA and ADD (UD230017TD).

\bibliography{aaai2026}

\clearpage
\setcounter{page}{1}

\supplementarytitle{Leveraging Textual Compositional Reasoning for Robust Change Captioning}

\setcounter{table}{0}
\renewcommand{\thetable}{S.\arabic{table}}

\setcounter{figure}{0}
\renewcommand{\thefigure}{S.\arabic{figure}}

\setcounter{equation}{0}
\renewcommand{\theequation}{S.\arabic{equation}}

\appendix

In this supplementary material, we provide an in-depth exploration of the various components and experimental analyses that complement the main paper as follows:

\begin{tcolorbox}[colback=gray!5!white, colframe=gray!75!black, title=Table of Contents]
\begin{enumerate}[label=\arabic*., leftmargin=*, itemsep=2pt]
  \item Detailed Description of Relation-aware Text Embedded (RTE) Dataset \dotfill \pageref{sec:suple_a}
  \item Detailed Prompts for Direct Input of Paired Images into Vision Language Model \dotfill \pageref{sec:suple_b}
  \item Effect of the Hyperparameter ($\lambda$) \dotfill \pageref{sec:suple_d}
  \item Effect of the Text Encoder \dotfill \pageref{sec:suple_e}
  \item Qualitative Results \dotfill \pageref{sec:suple_f}
  \item Discussion on Mitigating VLM Dependency \dotfill \pageref{sec:suple_g}
  \item Human Evaluation of VLM-generated Compositional Reasoning Text \dotfill \pageref{sec:suple_h}
  \item Detailed Error Analysis \dotfill \pageref{sec:suple_i}
  \item Code Availability \dotfill \pageref{sec:code}
\end{enumerate}
\end{tcolorbox}

This document provides additional insights, detailed analyses, and additional visualization results for deeper understanding of our proposed method and its significant contributions to the change captioning task.

\begin{table*}[!t]
    \centering
    \begin{mybox2}

\textbf{\textit{EXAMPLES}:}  \\
1. there are people on the stairs now\\
2. there is a person walking now\\
3. the person is not there anymore\\
4. the person walking is no longer there\\
5. person sitting at table far left moved slightly\\
6. there is a group of people in between the two buildings\\
7. the people in the previous picture are gone\\
8. there is not a person near the red car\\
9. there are 2 people in the last one that were not in the first one\\
10. the white car is not there anymore\\
11. the grey car in the back is not there anymore\\
12. there white car by the truck is not there anymore\\
13. there is a car in the middle now\\
14. there is a black car behind the red car in the middle\\
15. there is a black car in the middle row missing that was next to a silver car\\
16. black car is parked in after image and still driving in before image\\
17. there is less tables\\
18. shadow on umbrella at bottom left has changed a little bit\\
19. the before picture has a lady in front of the blue awning\\
20. the after picture contains two people walking towards the left\\
\\
\textbf{\textit{REQUIREMENTS}}: \\
1. Generate the caption describing the difference between the two images in ONE SINGLE SENTENCE ONLY.\\
2. DO NOT include any numbering, such as ``1.", ``2.", etc., or any bullet points.\\
3. DO NOT generate multiple sentences or paragraphs. The output MUST be one concise sentence.\\
4. If there are no changes, output must be ``there are no differences" or ``no change".\\
5. Recheck the output to confirm that it is ONLY ONE SENTENCE and follows the style of the EXAMPLES above.\\
    \end{mybox2}
     \caption{Guiding prompt for generating captions by describing changes between paired images using a Vision Language Model (VLM), with 20 GT caption examples provided to learn the style.}
     \label{prompt}
\end{table*}

\section{Detailed Description of Relation-aware Text Embedded (RTE) Dataset}
\label{sec:suple_a}
We introduce the Relation-aware Text Embedded (RTE) dataset, which consists of generated textual descriptions that explicitly capture compositional relationships between objects in the scene. We adopt the recent state-of-the-art VLM, InternVL \cite{internvl}, to generate RTE dataset.
Our RTE dataset can provide the enriched textual context for three widely-used datasets in the change captioning domain: CLEVR-Change \cite{DUDA}, CLEVR-DC \cite{clevrdc}, and Spot-the-Diff \cite{spotthediff} datasets. We refer to these augmented versions of the datasets as CLEVR-Change-RTE, CLEVR-DC-RTE, and Spot-the-Diff-RTE, respectively.

\begin{figure*}[t]
    \begin{minipage}[b]{1.0\linewidth}
	\centering
        \centerline{\includegraphics[width=0.99\linewidth]{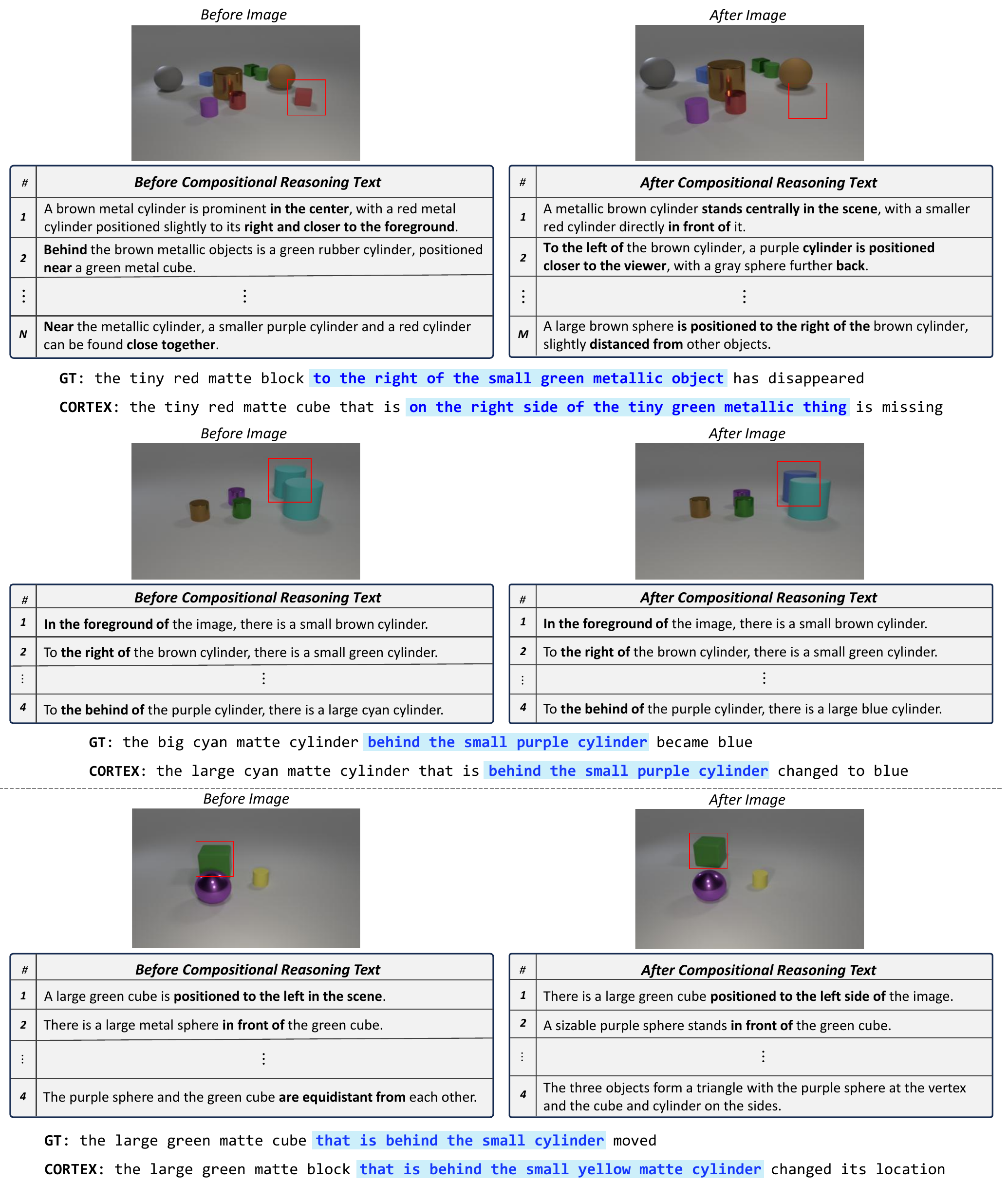}}
        \end{minipage}
    \centering
	\caption{Examples of CLEVR-Change-RTE dataset, which provides compositional reasoning text before and after image, with both ground-truth (GT) and predictions generated by CORTEX.}
    \label{fig:rte_change}
\end{figure*}

\begin{figure*}[t]
    \begin{minipage}[b]{1.0\linewidth}
	\centering
        \centerline{\includegraphics[width=0.97\linewidth]{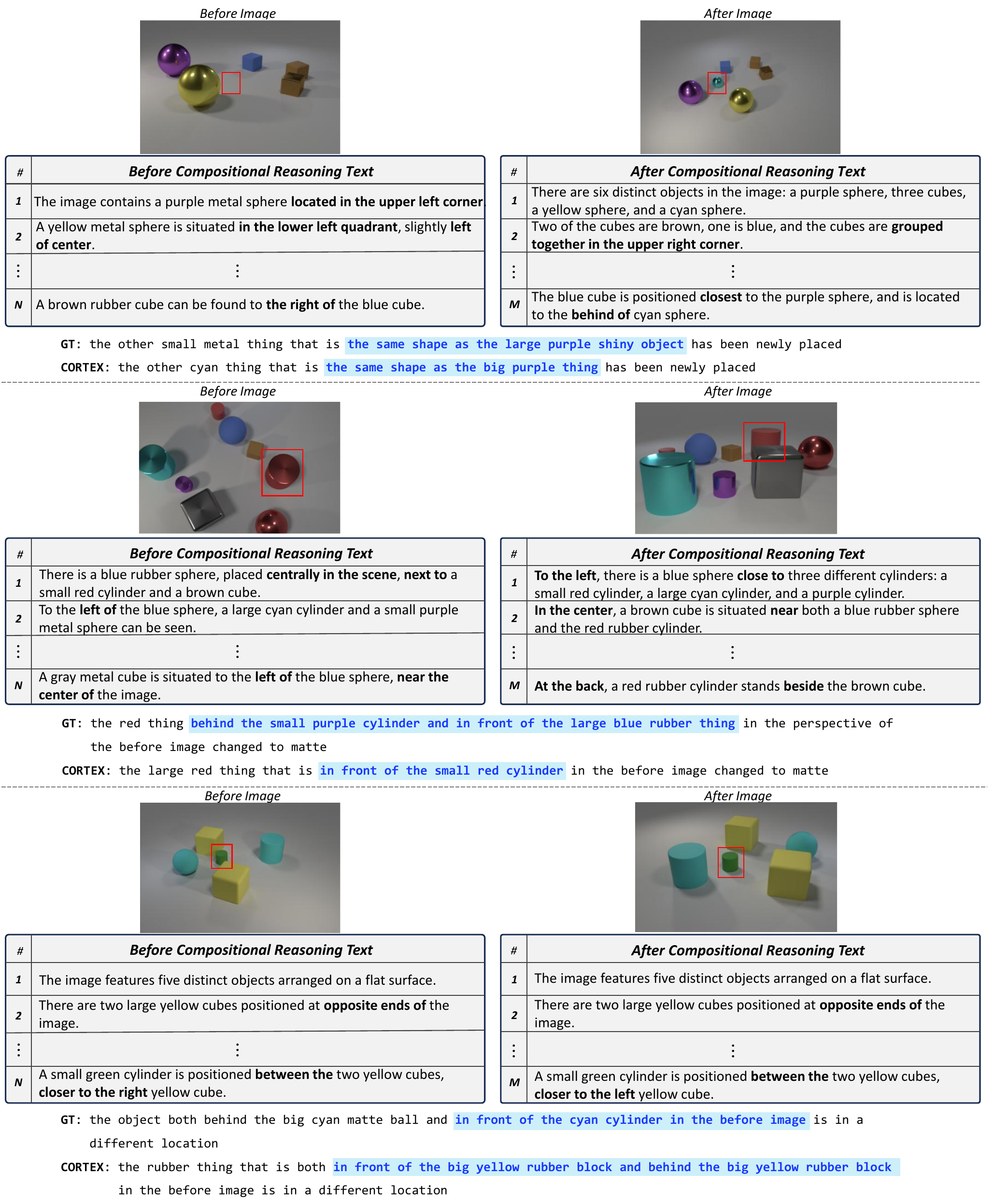}}
        \end{minipage}
    \centering
	\caption{Examples of CLEVR-DC-RTE dataset, which provides compositional reasoning text before and after image, with both ground-truth (GT) and predictions generated by CORTEX.}
    \label{fig:rte_dc}
\end{figure*}

\begin{figure*}[t]
    \begin{minipage}[b]{1.0\linewidth}
	\centering
        \centerline{\includegraphics[width=\linewidth]{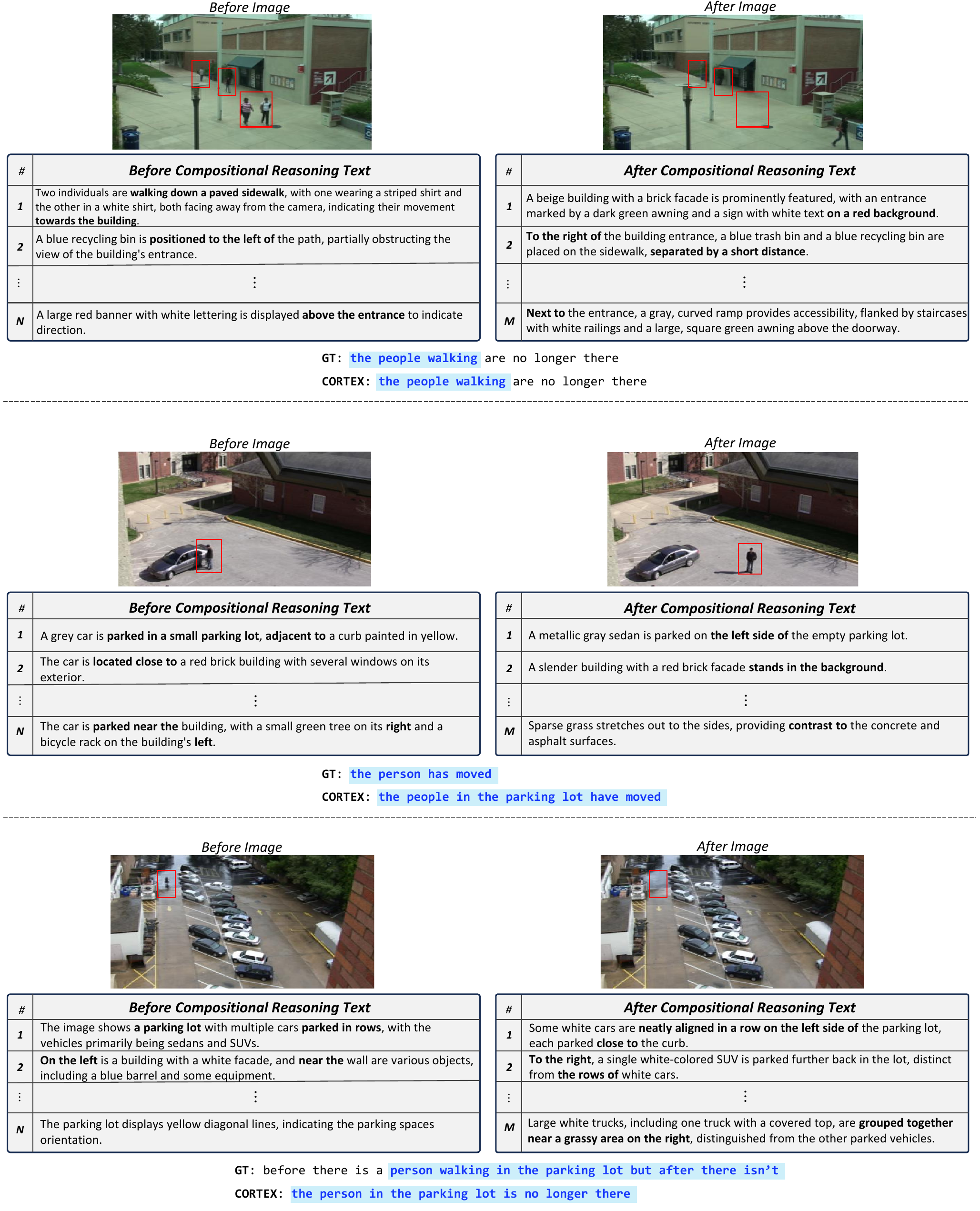}}
        \end{minipage}
    \centering
	\caption{Examples of Spot-the-Diff-RTE dataset, which provides compositional reasoning text before and after image, with both ground-truth (GT) and predictions generated by CORTEX.}
    \label{fig:rte_spot}
\end{figure*}

Figures \ref{fig:rte_change}, \ref{fig:rte_dc}, and \ref{fig:rte_spot} provide visual examples of the RTE datasets for CLEVR-Change-RTE, CLEVR-DC-RTE, and Spot-the-Diff-RTE, respectively. The RTE dataset consists of $N$ compositional reasoning sentences for the \textit{``before"} scene and $M$ for the \textit{``after"} scene. 
As mentioned in the main paper, $N$ and $M$ are determined through prompt tuning, enabling the VLM to autonomously analyze each scene and select the optimal number of sentences per image.
Each compositional reasoning sentence contains specific compositional reasoning cues, such as relative attributes (\textit{e.g.,} comparisons of size or brightness) and spatial relations, ensuring a detailed breakdown of scene components. This structure enables model to focus on compositional reasoning within the images, fostering a more comprehensive understanding of changes. 

Note that, in the Figures \ref{fig:rte_change}, \ref{fig:rte_dc}, and \ref{fig:rte_spot}, the lower section illustrates the ground truth (GT) captions and the result generated captions by our CORTEX. The compositional reasoning elements in the captions predicted by CORTEX, which perfectly match the GT captions, are highlighted in \hlblue{blue} to visualize the alignment.

\subsection{CLEVR-Change-RTE Dataset} Figure \ref{fig:rte_change} visualize an example from the CLEVR-Change-RTE dataset. Each scene is accompanied by individual sentences, each focusing on a specific object attributes and its relations within the scene. CLEVR-Change-RTE demonstrates the effectiveness of incorporating textual information in synthetic scenarios where controlled changes occur across various attributes, including color, size, and spatial positioning. In this dataset, the maximum number of captions per image is 15.

\subsection{CLEVR-DC-RTE Dataset} Figure \ref{fig:rte_dc} illustrates an example from the CLEVR-DC-RTE dataset. This dataset extends the textual enhancement of CLEVR-Change to scenarios involving drastic viewpoint changes, which pose significant challenges for change detection. By embedding compositional reasoning cues into the text, CLEVR-DC-RTE provides a stable reference for understanding object arrangements and relative attributes despite extreme viewpoint variations. This feature makes it a valuable resource for evaluating model performance under challenging spatial conditions, where pure visual features often struggle to maintain consistency. In this dataset, the maximum number of captions per image is 13.

\subsection{Spot-the-Diff-RTE Dataset} Figure \ref{fig:rte_spot} presents an example from the Spot-the-Diff-RTE dataset, which focuses on real-world surveillance scenarios. This dataset incorporates textual descriptions that provide detailed reasoning context for each object within complex, natural scenes. By capturing compositional reasoning cues, the Spot-the-Diff-RTE dataset aids in distinguishing meaningful changes (\textit{e.g.,} an object being added or removed) from irrelevant variations caused by lighting or background noise. This textual grounding significantly improves the robustness and interpretability of change captioning models in real-world applications. In this dataset, the maximum number of captions per image is 16.

\subsection{Significance of the RTE Dataset} The RTE dataset is a significant contribution to the field of change captioning, as it bridges the gap between purely visual understanding and explicit compositional reasoning. By embedding compositional reasoning into text, the RTE dataset provides a structured and interpretable layer of information that complements visual features. This additional textual guidance enhances the ability of models to detect and describe complex changes accurately, especially in scenarios with viewpoint shifts or cluttered backgrounds. Moreover, the availability of RTE-augmented versions of popular datasets enables researchers to benchmark their methods more effectively and explore novel approaches that leverage multi-modal representations.

\begin{figure*}[t]
    \begin{minipage}[b]{1.0\linewidth}
	\centering
        \centerline{\includegraphics[width=0.96\linewidth]{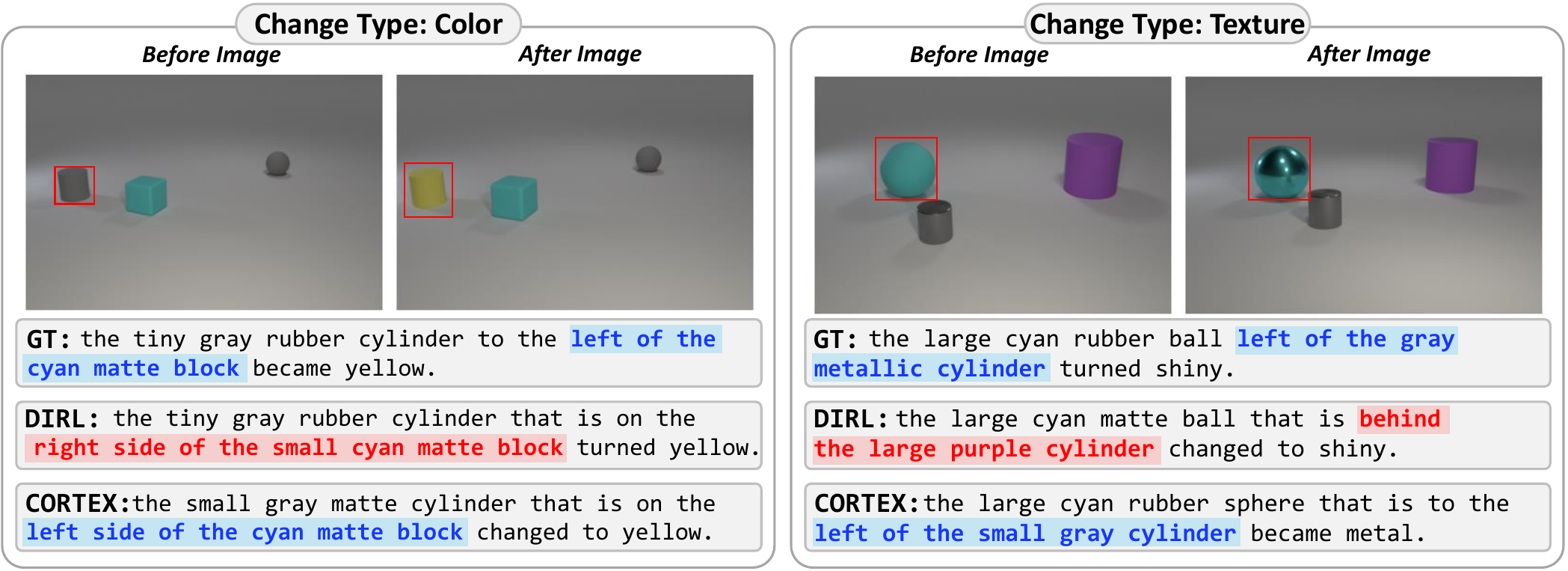}}
        \end{minipage}
    \centering
	\caption{Qualitative results of the CLEVR-Change dataset. Correct and incorrect predictions are highlighted using \hlblue{blue} and \hlred{red}, respectively.}
    \label{fig:q_change}
\end{figure*}

\begin{table*}[t] 
    \renewcommand{\tabcolsep}{3.5mm}
    \centering
    \resizebox{\linewidth}{!}
    {
    \begin{tabular}{c ccccc ccccc}
        \Xhline{3\arrayrulewidth}
        \multirow{2}{*}[-0.3em]{\textbf{Method}} & \multicolumn{5}{c}{\rule{0pt}{10pt}\bf Total Performance} & \multicolumn{5}{c}{\bf Semantic Change} \\
        \cmidrule(lr){2-6} \cmidrule(lr){7-11}
         & $\mathcal{B}$  & $\mathcal{M}$ & $\mathcal{R}$ & $\mathcal{C}$ & $\mathcal{S}$ & $\mathcal{B}$  & $\mathcal{M}$ & $\mathcal{R}$ & $\mathcal{C}$ & $\mathcal{S}$ \\
        \midrule
        Baseline (DIRL)
        &{55.5} & {40.8} & {73.4} & {125.3} & {33.4} & \textbf{55.4} & {38.4} & {72.1} & {123.2} & {32.7}
        \\\cdashline{1-11}
        \rule{0pt}{10.5pt}\bf CORTEX ($\lambda=10^{-1}$) 
        & \underline{57.0} & \underline{42.8} & \underline{76.0} & 130.2 & \textbf{34.3} & \underline{54.9} & \underline{39.3} & 74.2 & \underline{130.2} & \textbf{33.7} \\
        \bf CORTEX ($\lambda=10^{-2}$)
        & 56.2 & 41.7 & 75.2 & \underline{130.7} & \underline{34.2} & 54.6 & 39.1 & \underline{74.3} & \textbf{131.1} & 33.4 \\
        \bf CORTEX ($\lambda=10^{-3}$)
        & 56.3 & 41.2 & 75.2 & 130.5 & 33.9 & 54.7 & 38.4 & 74.2 & 130.0 & 32.7 \\
        \bf CORTEX ($\lambda=10^{-4}$)
        & \textbf{57.4} & \textbf{43.0} & \textbf{76.2} & \underline{130.7} & \underline{34.2} & \textbf{55.4} & \textbf{39.6} & \textbf{74.6} & \textbf{131.1} & \underline{33.5} \\
        \Xhline{3\arrayrulewidth}
    \end{tabular}
    }
    \caption{Performance comparison between the baseline visual-only method (DIRL) and our plug-and-play CORTEX with different $\lambda$ values on the CLEVR-Change dataset. The parameter $\lambda$ controls the alignment loss weight in the total objective function. Results show that CORTEX consistently improves performance over the baseline regardless of $\lambda$, with optimal results achieved at $\lambda=10^{-4}$. }
    \label{table:lambda}
\end{table*}

\section{Detailed Prompts for Direct Input of Paired Images into Vision Language Model}
\label{sec:suple_b}
We provide further details about the experiments described in subsection ``\textit{\textbf{Comparison of VLM Usage Strategies}}'' of ``\textbf{Experiments}'' section in the main paper, which compare different VLM usage strategies. In one experiment, the VLM is directly fed two images to generate a caption that describes the differences between them. This experiment was conducted on the Spot-the-Diff \cite{spotthediff} dataset, which reflects real-world scenarios. Paired images are simultaneously input into the VLM, and a carefully designed prompt is used to generate a caption that highlights the differences between the images.\\
As shown in Table \ref{prompt}, the prompt we used was constructed using an in-context learning approach. The \textit{\textbf{EXAMPLES}} contains 20 sentences taken from the ground truth (GT) of the Spot-the-Diff dataset, which helps familiarize the VLM with the GT caption style. The \textit{\textbf{REQUIREMENTS}} then directs the model to generate a caption that describes the differences between the two images in the same style as the GT captions.

While this direct inference approach provides a simple and intuitive way to utilize VLMs, it has inherent limitations. In particular, relying solely on VLM-generated text often leads to difficulty in capturing fine-grained or ambiguous differences, especially in complex scenarios involving multiple similar objects or changes in viewpoint. Since the textual descriptions are sparse and do not explicitly encode low-level visual variations, such cases tend to result in incomplete or misleading captions.

To address these challenges, our proposed CORTEX framework incorporates visual cues extracted from an image-level change detector, and aligns them with VLM-generated textual features. This dual-modality integration enables more robust scene understanding by compensating for the limitations of either modality alone. This allows CORTEX to perform more effective compositional reasoning and to generate change captions that are both more accurate and more detailed.

\section{Effect of the Hyper-parameter}
\label{sec:suple_d}
In this section, we investigate the impact of varying the hyper-parameter $\lambda$ on the performance of our CORTEX framework. Note that, $\lambda$ controls the weight of the alignment loss $\mathcal{L}_{align}$ in our total loss (Eq. (10)). To see the effect, we conducted experiments by varying $\lambda$ with values of $10^{-1}$, $10^{-2}$, $10^{-3}$, and $10^{-4}$ on the CLEVR-Change dataset \cite{DUDA}.

As shown in Table \ref{table:lambda}, our method outperforms most existing methods regardless of the variation in $\lambda$, highlighting the robustness of our approach. Specifically, the best performance is achieved when $\lambda$ is set to $10^{-4}$. Even with $\lambda$ values of $10^{-1}$, $10^{-2}$, and $10^{-3}$, CORTEX consistently outperforms most previous methods, emphasizing the effectiveness of incorporating reasoning text guidance. This demonstrates the adaptability of our framework to different hyperparameters while maintaining superior performance.

\section{Effect of the Text Encoder}
\label{sec:suple_e}
To investigate the effect of text encoders in CORTEX, we conducted experiments on the CLEVR-Change dataset by additionally introducing CLIP \cite{clip}, in addition to BERT \cite{bert} that was used as our main text encoder. As shown in Table \ref{table:dc7}, both text encoders outperformed the baseline, DIRL \cite{dirl}, which did not use text information. Also, BERT outperformed CLIP by demonstrating a deeper understanding of the sentences, which allowed BERT to better capture the compositional context.

\begin{figure*}[t]
    \begin{minipage}[b]{1.0\linewidth}
	\centering
        \centerline{\includegraphics[width=0.96\linewidth]{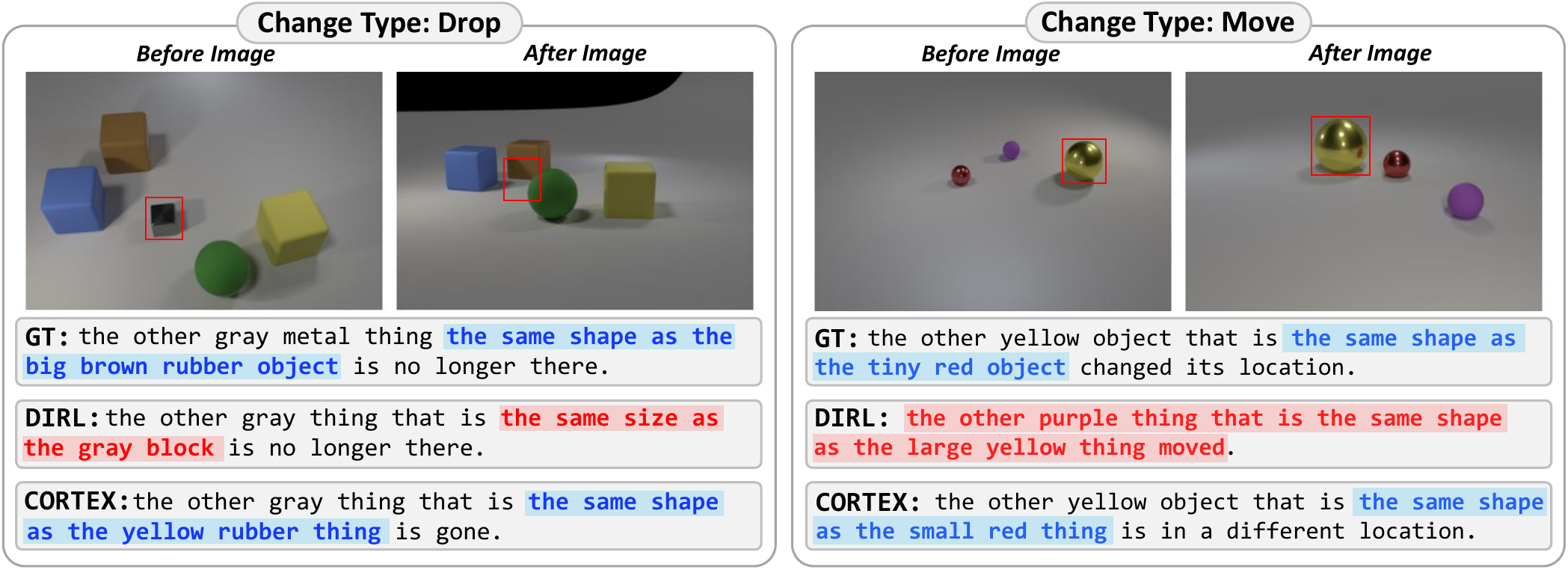}}
        \end{minipage}
    \centering
	\caption{Qualitative results of the CLEVR-DC dataset, which has moderate viewpoint change. Correct and incorrect predictions are highlighted using \hlblue{blue} and \hlred{red}, respectively.}
    \label{fig:q_dc1}
\end{figure*}

\begin{figure*}[t!]
    \begin{minipage}[b]{1.0\linewidth}
	\centering
        \centerline{\includegraphics[width=0.96\linewidth]{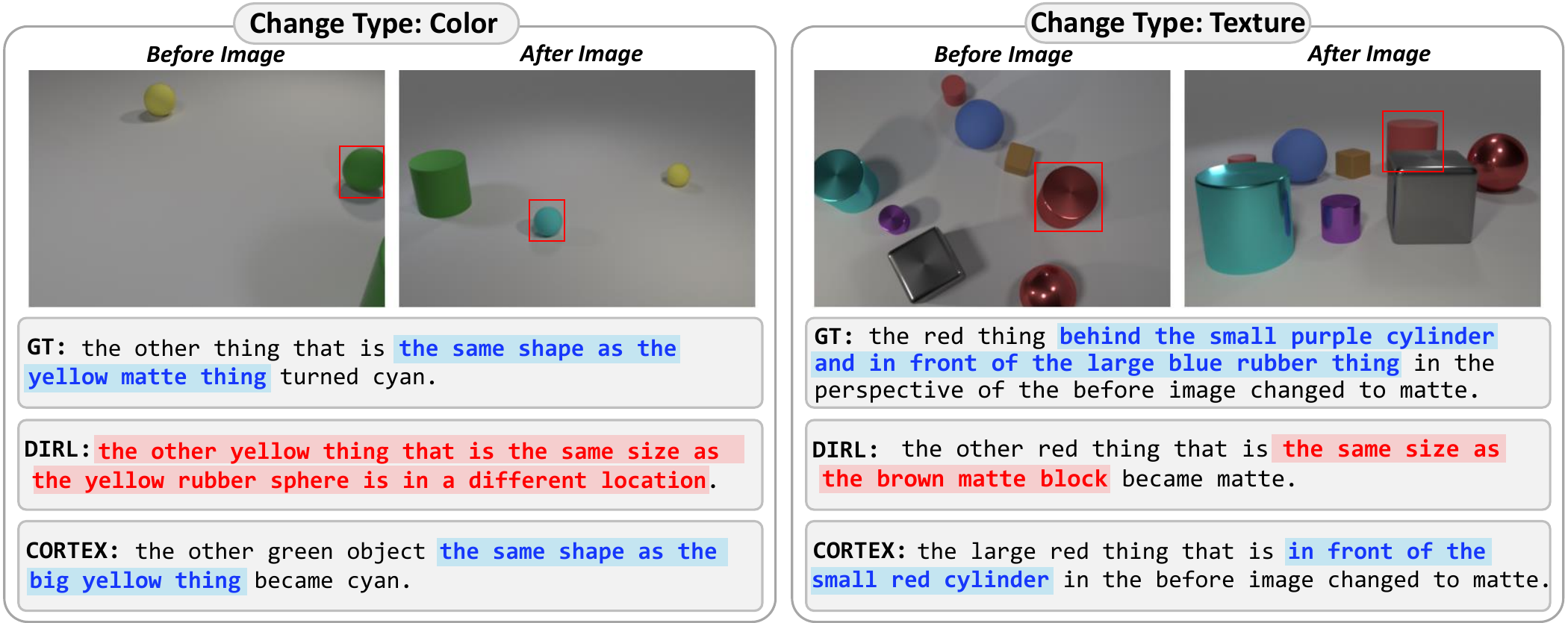}}
        \end{minipage}
    \centering
	\caption{Qualitative results of the CLEVR-DC dataset, which has drastic viewpoint change. Correct and incorrect predictions are highlighted using \hlblue{blue} and \hlred{red}, respectively.}
    \label{fig:q_dc2}
\end{figure*}
\begin{table}[t] 
    \renewcommand{\tabcolsep}{2.8mm}
    \centering
    \resizebox{\linewidth}{!}
    {
    \begin{tabular}{cccccc}
        \Xhline{3\arrayrulewidth}
        \rule{0pt}{10pt} \bf Text Encoder & $\mathcal{B}$  & $\mathcal{M}$ & $\mathcal{R}$ & $\mathcal{C}$ & $\mathcal{S}$ \\
        \midrule
        Baseline & 55.5 & 40.8 & 73.4 & 125.3 & 33.4  \\\cdashline{1-6}
        \rule{0pt}{11.5pt}CLIP & 55.5 & 41.3 & 75.1 & 128.3 & 33.6 \\
        BERT & \textbf{57.4} & \textbf{43.0} & \textbf{76.2} & \textbf{130.7} & \textbf{34.2} \\
        \Xhline{3\arrayrulewidth}
    \end{tabular}
    }
    \caption{Comparison of different text encoders (CLIP and BERT) in our method on the CLEVR-Change dataset. The top row represents the baseline DIRL \cite{dirl}, which is an visual-only method without text encoder.}
    \label{table:dc7}
\end{table}
\begin{figure*}[t]
    \begin{minipage}[b]{1.0\linewidth}
	\centering
        \centerline{\includegraphics[width=0.96\linewidth]{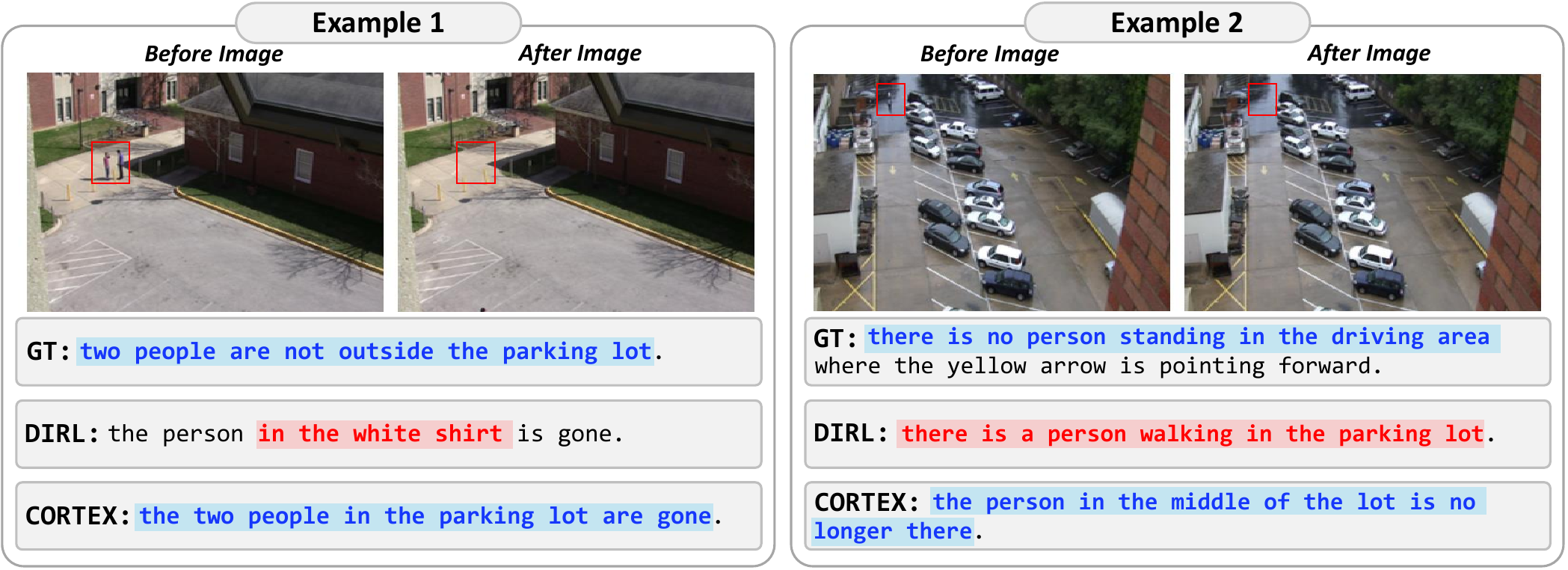}}
        \end{minipage}
    \centering
	\caption{Qualitative results of the Spot-the-Diff dataset. Correct and incorrect predictions are highlighted using \hlblue{blue} and \hlred{red}, respectively.}
    \label{fig:q_spot}
\end{figure*}

\section{Qualitative Results}
\label{sec:suple_f}
This section provides additional comparisons of the change captioning results between our CORTEX and the state-of-the-art method with publicly available source code, DIRL \cite{dirl}, which has the latest publicly available code. We present the results on the CLEVR-Change, CLEVR-DC, and Spot-the-Diff datasets. 

Figure \ref{fig:q_change} shows qualitative results from the CLEVR-Change dataset \cite{DUDA}, where CORTEX consistently outperforms DIRL \cite{dirl} in identifying fine-grained relative attributes and relationships, as well as in generating accurate change descriptions. These examples further validate the effectiveness of reasoning text guidance in improving compositional understanding.

Figure \ref{fig:q_dc1} and Figure \ref{fig:q_dc2} extend this comparison to the CLEVR-DC dataset \cite{clevrdc}, which includes various viewpoint changes. Even in these scenarios, our method successfully captures object relationships and accurately detects changes, demonstrating the ability to effectively handle extreme viewpoint shifts. This emphasizes a key strength: the capability to leverage compositional reasoning cues to adapt to drastic viewpoint changes. Unlike DIRL \cite{dirl}, which often struggles in such scenarios, CORTEX leverages its reasoning abilities to maintain strong performance, producing accurate and detailed captions.

Figure \ref{fig:q_spot} presents additional results from the Spot-the-Diff dataset \cite{spotthediff}, a real-world dataset with diverse and complex changes. Our method demonstrates strong generalization to natural scenes, outperforming DIRL by effectively identifying meaningful changes while filtering out irrelevant variations.

\section{Discussion on Mitigating VLM Dependency}
\label{sec:suple_g}

Although our method uses a Vision-Language Model (VLM) to generate compositional reasoning sentences, we mitigate potential scalability concerns through an offline preprocessing step to generate textual descriptions. To further reduce computational overhead, future studies could adopt knowledge distillation techniques, training a lightweight network to mimic the compositional reasoning capabilities of the original VLM. This distilled model can then perform inference efficiently without repeated heavy computations, significantly improving applicability of the model in large-scale, real-world scenarios.

\section{Human Evaluation of VLM-generated Compositional Reasoning Text}
\label{sec:suple_h}
Since the quality of VLM generated compositional reasoning sentences critically affects the performance of our CORTEX framework, we conducted a human evaluation to assess their reliability. A total of 60 image sets were randomly sampled, comprising 20 sets from each of the three datasets (CLEVR-Change \cite{DUDA}, CLEVR-DC \cite{clevrdc}, and Spot-the-Diff \cite{spotthediff}). 25 independent annotators participated in the evaluation, including both individuals with relevant domain expertise and individuals without prior knowledge. No example sentences were shared in advance, ensuring a fair assessment based solely on the naturalness and appropriateness of the sentences from a human perspective.

Each VLM generated sentence was evaluated according to three criteria: Accuracy, which measures whether the sentence correctly describes object attributes (\textit{e.g.,} color, shape, and material) and spatial relationships; Relevance, which assesses whether the described relational information is appropriate and meaningful for the given scene; and Fluency, which evaluates grammatical correctness and the natural flow of the sentence. Each criterion was scored on a scale from \textbf{1 (poor) to 5 (excellent)}, and the average scores across the 25 annotators are reported in Table~S.4.

\begin{table}[t]
\centering
\resizebox{0.9\linewidth}{!}{
\begin{tabular}{lccc}
\toprule
\textbf{Dataset} & \textbf{Accuracy} & \textbf{Relevance} & \textbf{Fluency} \\
\midrule
CLEVR-Change & 4.05 & 3.78 & 4.41 \\
CLEVR-DC & 4.22 & 3.98 & 4.51 \\
Spot-the-Diff & 3.90 & 3.78 & 4.34 \\
\bottomrule
\end{tabular}}
\caption{Human evaluation results for VLM-generated compositional reasoning sentences. Scores range from 1 (poor) to 5 (excellent) and reflect three aspects: Accuracy, relevance, and fluency.}
\label{table:human_eval_vlm}
\end{table}

\begin{itemize}
    \item \textbf{Accuracy:} Does the sentence correctly describe objects' attributes (\textit{e.g.,} color, shape, material) and spatial relationships?
    \item \textbf{Relevance:} Does the described relational information accurately reflect the relationships between objects within the image (\textit{e.g.,} size, position)
    \item \textbf{Fluency:} Is the sentence grammatically correct and natural?
\end{itemize}

Building on these results, we observe consistently strong scores on the two CLEVR datasets, while Spot-the-Diff dataset shows slightly lower Accuracy and Relevance, likely due to its more diverse scenes, clutter, and lighting variation. Nevertheless, the overall evaluation confirms that these sentences provide a reliable and high-quality textual basis for multimodal reasoning process of CORTEX.

\section{Detailed Error Analysis}
\label{sec:suple_i}
We analyzed the captions generated by the VLM within CORTEX to identify common error types present in the extracted compositional reasoning sentences.  
From 200 sampled erroneous sentences, we categorized errors into three types:  

\begin{enumerate}[label=(\arabic*)]
    \item \textbf{Spatial Relation Errors:} Incorrect understanding of spatial relations (\textit{e.g.,} left-right or front-back misinterpretations).
    \item \textbf{Attribute Identification Errors:} Incorrect descriptions regarding object attributes (\textit{e.g.,} wrong color, size, or shape).
    \item \textbf{Missing or Extra Object Errors:} Errors related to incorrectly identifying objects that appear or disappear between scenes.
\end{enumerate}

Examples of these error types are illustrated in Figure~S.8 (CLEVR-Change \cite{DUDA}), Figure~S.9 (CLEVR-DC \cite{clevrdc}), and Figure~S.10 (Spot-the-Diff \cite{spotthediff}).  
Each figure shows representative cases where captions extracted from images contain such errors, along with the corresponding ground truth (GT) captions and the final captions generated by CORTEX.

While VLM-generated captions can contain the above three types of errors, our method mitigates them through two key mechanisms:  
First, instead of extracting only a single caption per scene from the VLM, we generate multiple dynamic captions for each scene. These captions provide complementary perspectives, allowing the model to form a deeper and more robust understanding of the scene.  
Second, CORTEX incorporates visual cues extracted from the image-level change detector, which encode visual differences between the ``before'' and ``after'' images. By integrating both the textual modality (compositional reasoning sentences) and the visual modality (image-level change detected features), our framework performs a more comprehensive scene understanding. This multimodal fusion enables more precise compositional reasoning and ultimately leads to the generation of more specific and accurate change captions.

\section{Code Availability}
\label{sec:code}
To support reproducibility, we have uploaded the implementation of our model architecture, including key components, as part of the supplementary materials during the review process. The full source code, including training and evaluation scripts, will be made publicly available after the review is complete.


\begin{figure*}[t]
    \begin{minipage}[b]{0.7\linewidth}
	\centering
        \centerline{\includegraphics[width=0.99\linewidth]{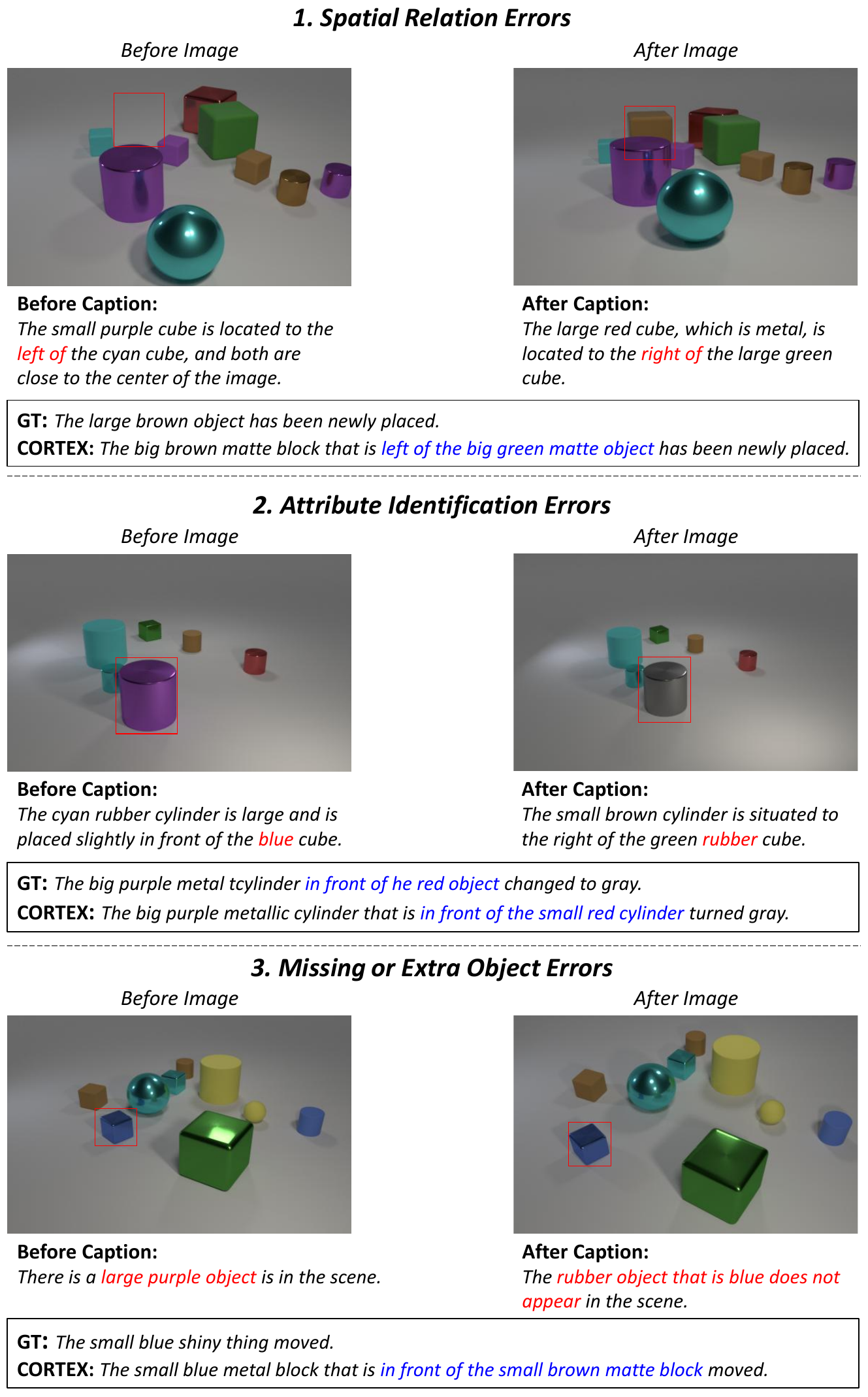}}
        \end{minipage}
    \centering
	\caption{Examples of error analysis on the CLEVR-Change dataset.}
    \label{fig:error_chg}
\end{figure*}

\begin{figure*}[t]
    \begin{minipage}[b]{0.7\linewidth}
	\centering
        \centerline{\includegraphics[width=0.99\linewidth]{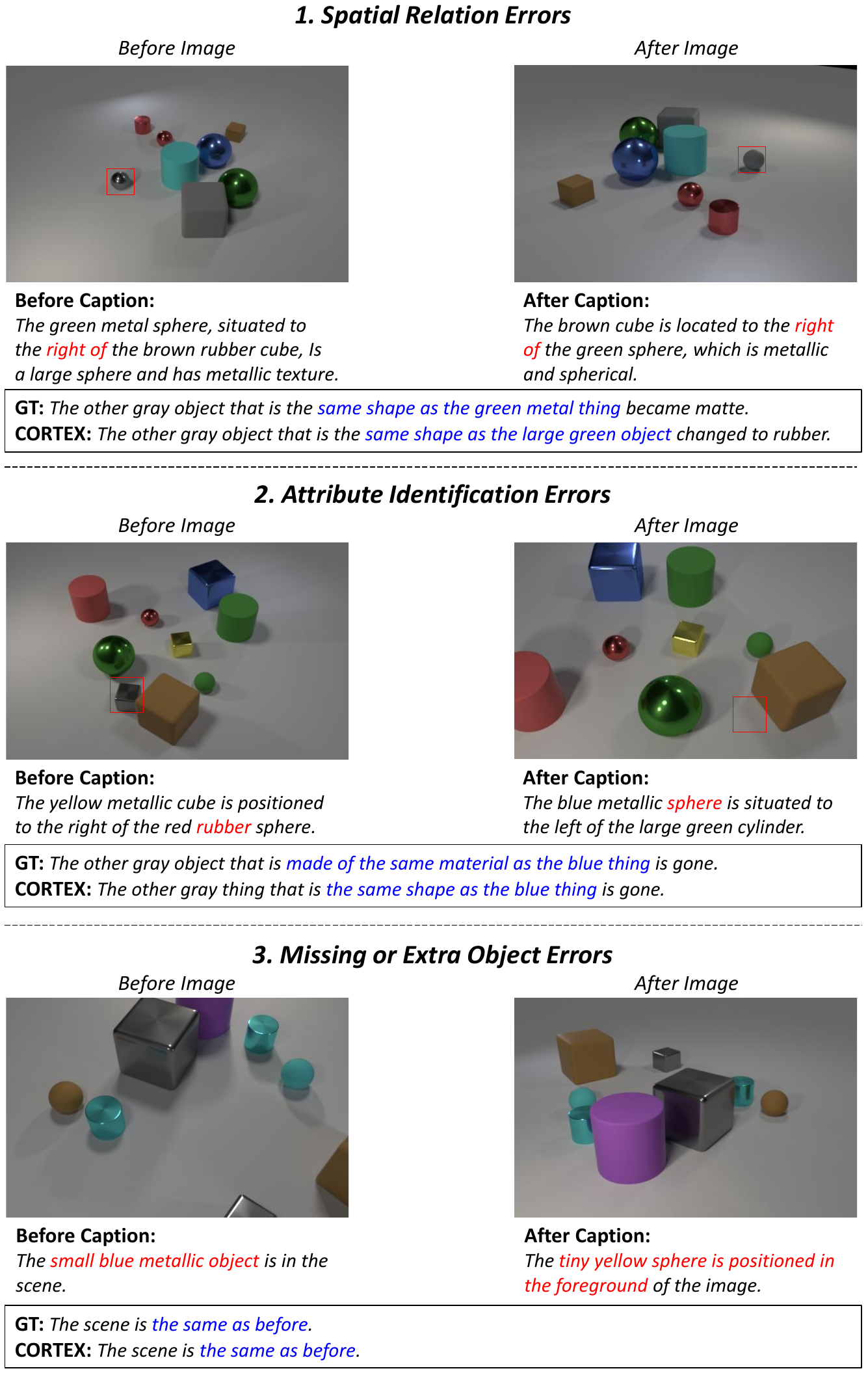}}
        \end{minipage}
    \centering
	\caption{Examples of error analysis on the CLEVR-DC dataset.}
    \label{fig:error_dc}
\end{figure*}

\begin{figure*}[t]
    \begin{minipage}[b]{0.7\linewidth}
	\centering
        \centerline{\includegraphics[width=0.99\linewidth]{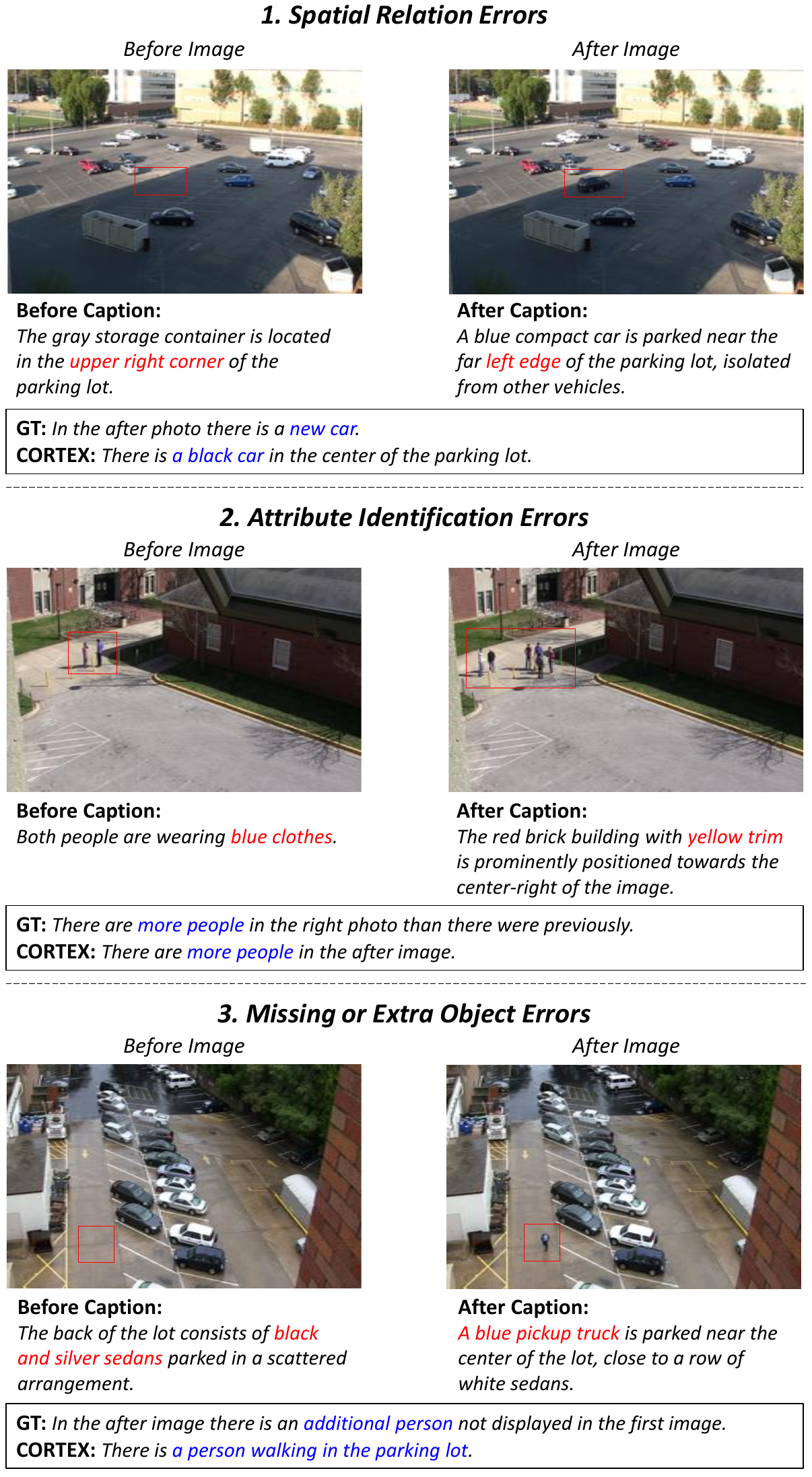}}
        \end{minipage}
    \centering
	\caption{Examples of error analysis on the Spot-the-Diff dataset.}
    \label{fig:error_std}
\end{figure*}

\section*{References}
\small Chen, Z.; Wu, J.; Wang, W.; Su, W.; Chen, G.; Xing, S.; Zhong, M.; Zhang, Q.; Zhu, X.; Lu, L.; et al. 2024. Internvl: Scaling up vision foundation models and aligning for generic visual-linguistic tasks. In Proceedings of the IEEE/CVF Conference on Computer Vision and Pattern Recognition, 24185–24198. \\
\small Devlin, J.; Chang, M.-W.; Lee, K.; and Toutanova, K. 2019. BERT: Pre-training of Deep Bidirectional Transformers for Language Understanding. arXiv:1810.04805. \\
\small Jhamtani, H.; and Berg-Kirkpatrick, T. 2018. Learning to describe differences between pairs of similar images. arXiv preprint arXiv:1808.10584. \\
\small Kim, H.; Kim, J.; Lee, H.; Park, H.; and Kim, G. 2021. Viewpoint-Agnostic Change Captioning with Cycle Consistency. In ICCV. \\
\small Park, D. H.; Darrell, T.; and Rohrbach, A. 2019. Robust change captioning. In Proceedings of the IEEE/CVF International Conference on Computer Vision, 4624–4633. \\
\small Radford, A.; Kim, J. W.; Hallacy, C.; Ramesh, A.; Goh, G.; Agarwal, S.; Sastry, G.; Askell, A.; Mishkin, P.; Clark, J.; et al. 2021. Learning transferable visual models from natural language supervision. In International conference on machine learning, 8748–8763. PMLR. \\
\small Tu, Y.; Li, L.; Su, L.; Yan, C.; and Huang, Q. 2024. Distractors-immune representation learning with crossmodal contrastive regularization for change captioning. In European Conference on Computer Vision, 311–328. Springer. \\

\end{document}